\documentclass[journal]{IEEEtran}
\usepackage{bbm}
\usepackage{amssymb}
\usepackage{amsmath}
\usepackage{tikz}
\usepackage{forest}

\usetikzlibrary{positioning, decorations.pathreplacing}

\usepackage{hyperref}
\usepackage{cleveref}

\ifCLASSINFOpdf
\else
\fi
\usepackage{xcolor}

\usepackage{authblk}

\title{A Comprehensive Survey of Mixture-of-Experts: Algorithms, Theory, and Applications}

\author{Siyuan Mu and Sen Lin
\IEEEcompsocitemizethanks{
\IEEEcompsocthanksitem The work Siyuan Mu was done when she was an intern at University of Houston. (e-mail: mu13312316567@gmail.com). 
\IEEEcompsocthanksitem Sen Lin is with the Department of Computer Science, University of Houston, Houston, TX, USA (e-mail: slin50@central.uh.edu).

\IEEEcompsocthanksitem This work has been submitted to the IEEE for possible publication. Copyright may be transferred without notice, after which this version may no longer be accessible. }
}

\begin{document}

\maketitle

\begin{abstract}
Artificial intelligence (AI) has achieved astonishing successes in many domains, especially with the recent breakthroughs in the development of foundational large models. These large models, leveraging their extensive training data, provide versatile solutions for a wide range of downstream tasks. However, as modern datasets become increasingly diverse and complex, the development of large AI models faces two major challenges: (1) the enormous consumption of computational resources and deployment difficulties, and (2) the difficulty in fitting heterogeneous and complex data, which limits the usability of the models. Mixture of Experts (MoE) models has recently attracted much attention in addressing these challenges, by dynamically selecting and activating the most relevant sub-models to process input data. It has been shown that MoEs can significantly improve model performance and efficiency with fewer resources, particularly excelling in handling large-scale, multimodal data. Given the tremendous potential MoE has demonstrated across various domains, it is urgent to provide a comprehensive summary of recent advancements of MoEs in many  important fields. Existing surveys on MoE have their limitations, e.g., being outdated or lacking discussion on certain key areas, and we aim to address these gaps. In this paper, we first introduce the basic design of MoE, including gating functions, expert networks, routing mechanisms, training strategies, and system design. We then explore the algorithm design of MoE in important machine learning paradigms such as continual learning, meta-learning, multi-task learning, reinforcement learning, and federated learning. Additionally, we summarize theoretical studies aimed at understanding MoE and review its applications in computer vision and natural language processing. Finally, we discuss promising future research directions.
\end{abstract}
\section{Introduction}

Artificial Intelligence (AI) has emerged as a transformative force in modern technologies, demonstrating  remarkable successes  across multiple domains from computer vision \cite{dosovitskiy2020image,kirillov2023segment,shi2023yolov} and speech recognition \cite{radford2023robust,an2024funaudiollm} to healthcare \cite{rajpurkar2022ai} and autonomous systems \cite{cao2022autoai}. A key driver behind this is the development of foundation models, such as BERT \cite{devlin2019bert}, CLIP \cite{radford2021learning}, GPT-4 \cite{achiam2023gpt} which are large-scale neural networks pre-trained on massive datasets and  provide versatile solutions for a wide range of downstream tasks. Building upon the pre-trained knowledge, these models can be further fine-tuned with minimal additional training to adapt to specific applications, making them indispensable tools for handling increasingly complex and diverse tasks in many real-world scenarios.

However, as AI applications expand, modern datasets are becoming more diverse and complex. They often contain multimodal data (e.g., text, images, and audio) and exhibit intricate structures (e.g., graphs or hierarchical relationships).  This diversity and complexity introduce two significant challenges: (1) the computational cost of training and deploying large models grows exponentially, making it unsustainable for many applications \cite{thompson2020computational}, and (2) integrating conflicting or heterogeneous knowledge within a single model becomes increasingly difficult, often leading to unstable training dynamics and suboptimal performance \cite{sun2017revisiting}. These challenges underscore the need for more efficient and scalable architectures capable of meeting the growing demands of modern AI tasks.

One promising approach to addressing these challenges is the Mixture of Experts (MoE) architecture, which has attracted much attention recently. Originally proposed in \cite{jacobs1991adaptive,jordan1994hierarchical}, MoE adopts a ``divide and conquer" strategy that fundamentally differs from traditional dense models. While conventional models activate all parameters for every input, MoE models  dynamically select and activate only the most relevant subset of parameters based on the characteristics of the input data. This approach not only enhances the specialization of individual experts but also mitigates the difficulties associated with training on diverse and conflicting tasks. Furthermore, the selective activation mechanism allows MoE models to significantly expand their capacity and handle diverse knowledge domains without proportionally increasing computational costs, thereby achieving an optimal balance between performance and efficiency \cite{fedus2022switch,shazeer2017outrageously}. By leveraging the strengths of specialized ``experts" for different tasks or data types, MoE provides a scalable and flexible framework for tackling the challenges posed by complex, multifaceted datasets.

\definecolor{line-color}{RGB}{0, 119, 255}
\definecolor{fill-color}{RGB}{218, 191, 216}

\tikzstyle{category}=[
    rectangle,
    draw=line-color,
    rounded corners,
    text opacity=1,
    minimum height=1.5em,
    minimum width=5em,
    inner sep=2pt,
    align=center,
    fill opacity=.5,
]
\tikzstyle{leaf}=[category,minimum height=1.5em,
fill=fill-color!40, text width=20em,  text=black,align=left,font=\scriptsize,
inner xsep=2pt,
inner ysep=1pt,
]

\begin{figure*}[tp]
  \centering

\begin{forest} for tree={
  grow=east,
  reversed=true,
  anchor=base west,
  transform shape,   
  scale=1.1,       
  parent anchor=east,
  child anchor=west,
  base=left,
  font=\small,
  rectangle,
  draw=line-color,
  rounded corners,align=left,
  minimum width=4.0em,
  s sep=4pt,
  inner xsep=4pt,
  inner ysep=4pt,
  align=left,
  ver/.style={rotate=90, child anchor=north, parent anchor=south, anchor=center},    
},
  where level=1{text width=5.5em,font=\scriptsize,}{},
  where level=2{text width=5.5em,font=\scriptsize}{},
  where level=3{text width=5.5em,font=\scriptsize}{},
  where level=4{text width=5.5em,font=\scriptsize}{},
 [Mixture of Experts (MoE), ver
    [Basics of MoE
        [Gating Function
            [Switch transformers \cite{fedus2022switch}{,} V-MoE \cite{riquelme2021scaling}{,} RMoE \cite{wu2022residual}{,} M3vit \cite{fan2022m3vit}{,} GLaM \cite{du2022glam}{,} 
            Shazeer \textit{et al.} \cite{shazeer2017outrageously}{,}\\  Gshard \cite{lepikhin2020gshard}{,} Mixtral of experts \cite{jiang2024mixtral}{,} GMoE \cite{li2022sparse}{,} Chi \textit{et al.} \cite{chi2022representation}{,} Nguyen \textit{et al.} \cite{nguyen2024statistical}{,} Xu \textit{et al.} \cite{xu1994alternative}{,}\\ softMoE \cite{puigcerver2023sparse}{,} Geweke \textit{et al.} \cite{geweke2007smoothly},leaf,text width=30em]
        ]
        [Expert Network
            [Xu \textit{et al.} \cite{xu1994alternative}{,} softMoE \cite{puigcerver2023sparse}{,} Switch transformers \cite{fedus2022switch}{,} V-MoE \cite{riquelme2021scaling}{,} Shazeer \textit{et al.} \cite{shazeer2017outrageously}{,}
            Gshard \cite{lepikhin2020gshard}{,}\\ Chen \textit{et al.} \cite{chen2022towards}{,} Li \textit{et al.} \cite{li2024theory1}{,} Switchhead \cite{csordas2023switchhead}{,}MoA \cite{wang2024moa} MoH \cite{jin2024moh}{,} pMoE \cite{jung2024pmoe}{,} Pavlitska \textit{et al.} \cite{pavlitska_sparsely-gated_2023}{,}\\Yi \textit{et al.} \cite{yi_mixture_2019}{,} Zhang \textit{et al.} \cite{zhang_enhancing_2022}{,} 
            Zhang \textit{et al.} \cite{zhang_learning_2019}{,} Gross \textit{et al.} \cite{gross2017hard},leaf,text width=30em]
        ]
        [Routing Strategy
            [V-MoE \cite{riquelme2021scaling}{,} pMoE \cite{jung2024pmoe}{,} Uni-perceiver-moe \cite{zhu2022uni}{,} Maskmoe \cite{su2024maskmoe}{,} Pedicir \textit{et al.} \cite{pedicir2024novel}{,} Uni-moe \cite{li2024uni}{,}\\ Kudugunta \textit{et al.} \cite{kudugunta2021beyond} Yi \textit{et al.} \cite{yi_mixture_2019} {,} Shi \textit{et al.} \cite{shi2024time},leaf,text width=30em]
        ]
        [Training Strategy
            [Switch transformers \cite{fedus2022switch}{,} V-MoE \cite{riquelme2021scaling}{,} Shazeer \textit{et al.} \cite{shazeer2017outrageously}{,} RMoE \cite{wu2022residual}{,} Huang \textit{et al.} \cite{huang2024harder}{,} Hmoe \cite{wang2024hmoe}{,}\\Irsoy \textit{et al.} \cite{irsoy2021dropout}{,} Faster-moe \cite{he2022fastermoe},leaf,text width=30em]
        ]
        [System Design
            [Switch transformers \cite{fedus2022switch}{,} M3vit \cite{fan2022m3vit}{,} Tutel \cite{hwang2023tutel}{,} softMoE \cite{puigcerver2023sparse}{,}  Faster-moe \cite{he2022fastermoe}{,} Edge-moe \cite{sarkar2023edge}{,}\\ Deepspeed-MoE \cite{rajbhandari2022deepspeed}{,}DeepSeek-V3 \cite{liu2024deepseek}{,} Singh \textit{et al.} \cite{singh2023hybrid}{,} Yao \textit{et al.} \cite{yao2024exploiting},leaf,text width=30em]
        ]
    ]
    [Algorithms
        [Continual \\ Learning
            [Yu \textit{et al.} \cite{Yu_2024_CVPR}{,} Zhang \textit{et al.} \cite{zhang2024decomposing}{,} Zhou \textit{et al.} \cite{zhou2022mixture}{,} Lifelong-MoE \cite{chen2023lifelong}{,} Lee \textit{et al.} \cite{lee2020neural}{,} SEED \cite{rypesc2024divide}{,} \\Evolve \cite{CL_9}{,} Mote \cite{li5035279mote}{,} Park \textit{et al.} \cite{park2024learning}{,}Li \textit{et al.} \cite{li2024theory1}{,} Hihn \textit{et al.} \cite{hihn2021mixture}{,} Le \textit{et al.} \cite{le2024mixture}{,} Lee \textit{et al.} \cite{lee2024continual}{,}\\ PMoE \cite{jung2024pmoe}{,} Wang \textit{et al.} \cite{CL_19}{,} Chen \textit{et al.} \cite{CL_20}{,} Wang \textit{et al.} \cite{wang2022learning} {,} Aljundi \textit{et al.} \cite{aljundi2017expert} {,}  Wang \textit{et al.} \cite{wang2022coscl}{,}\\ Doan \textit{et al.} \cite{doan2023continual}
            ,leaf,text width=30em]
        ]
        [Meta \\ Learning
                [Meta-DMoE \cite{zhong2022meta}{,} MoE-NPs \cite{wang2022learning}{,} RaMoE \cite{MeL_3}{,} Liu \textit{et al.} \cite{liu2024meta}{,} Guo \textit{et al.} \cite{guo2018multi}{,} Zhou \textit{et al.} \cite{MeL_6}{,}\\ MixER \cite{nzoyem2025towards}
                ,leaf,text width=30em]
        ]
        [Multi-task \\ Learning
            [MOOR \cite{hendawy2023multi}{,} Park \textit{et al.} \cite{park2024learning}{,} MLoRE \cite{yang2024multi}{,} WEMoE \cite{shen2024efficient}{,} MoSE \cite{MTL_7}{,} MMoEEx \cite{MTL_10}{,} Chen \textit{et al.} \cite{cheng2023multi}{,} \\ TaskExpert \cite{Ye_2023_ICCV}{,} Gupta \textit{et al.} \cite{gupta2022sparsely}{,} DSelect-k \cite{hazimeh2021dselect}{,}  Mod-Squad \cite{MTL_16}{,} MoDE \cite{reuss2024efficient}{,} M3oE \cite{MTL_9}{,}\\ MoME \cite{MTL_6}{,} 
            TI-Expert \cite{MTL_4}{,} Ma \textit{et al.} \cite{MTL_18}{,} Hou and Cao \textit{et al.} \cite{MTL_20}{,} CMoIE \cite{MTL_21}{,} Sodhani \textit{et al.} \cite{sodhani2021multi}{,} \\AdaMV-MoE \cite{MTL_12}{,} Tang \textit{et al.} \cite{tang2020progressive}{,} M$^3$vit \textit{et al.} \cite{fan2022m3vit}{,} Louizos \textit{et al.} \cite{louizos2017learning}{,} Jacobs \textit{et al.} \cite{jacobs1991adaptive}
            ,leaf,text width=30em
            ]
        ]
        [Reinforcement \\ Learning
            [Ren \textit{et al.} \cite{ren2021probabilistic}{,} Gimelfarb \textit{et al.} \cite{pmlr-v161-gimelfarb21a}{,} Willi \textit{et al.} \cite{willi2024mixture}{,} MENTOR \cite{huang2024mentor}{,} MMICRL \cite{qiao2023multi}{,} \\Gupta  \textit{et al.} \cite{gupta2023offline}{,}  Samejima \textit{et al.} \cite{samejima2003inter}{,} Van \textit{et al.} \cite{van2008switching}{,} MACE \cite{peng2016terrain}{,} Kumar \textit{et al.} \cite{kumar2017learning} {,} \\ Peng \textit{et al.} \cite{peng2019mcp}{,} Akrour \textit{et al.} \cite{akrour2021continuous}{,} MVE \cite{refaat2021accelerated}{,} Germ \cite{song2024germ}{,} SMoSE \cite{vincze2024smose}{,} Obando \textit{et al.} \cite{obando2024mixtures}{,}\\ Takahashi \textit{et al.} \cite{takahashi2004modular}{,}Takahashi \textit{et al.} \cite{takahashi2005modular}{,} Takahashi \textit{et al.} \cite{takahashi2005simultaneous}{,} Mulling \textit{et al.} \cite{mulling2013learning}{,} Li \textit{et al.} \cite{li2024mixtures}{,}\\ Ewerton \textit{et al.} \cite{ewerton2015learning}{,} Zhou \textit{et al.} \cite{zhou2020movement}{,} Freymuth \textit{et al.} \cite{freymuth2022inferring}{,} Prasad \textit{et al.} \cite{prasad2024moveint}{,} MMRL \cite{doya2002multiple}{,}\\TERL \cite{tommasino2016reinforcement} {,}  ,leaf,text width=30em]
        ]
        [Federated \\ Learning
            [Peterson \textit{et al.} \cite{peterson2019private}{,} Zec \textit{et al.} \cite{zec2020specialized}{,} Pye \textit{et al.} \cite{pye2021personalised}{,} Reisser \textit{et al.} \cite{reisser2021federated}{,} Guo \textit{et al.} \cite{guo2021pfl}{,} \\ Isaksson \textit{et al.} \cite{isaksson2022adaptive}{,} Ghosh \textit{et al.} \cite{ghosh2020efficient}{,} Tran \textit{et al.} \cite{tran2025revisiting}{,} Dun \textit{et al.} \cite{dun2023fedjets}{,} Heinbaught \textit{et al.} \cite{heinbaugh2023data}{,}\\ Su \textit{et al.} \cite{su2023one}{,} Zeng \textit{et al.} \cite{zeng2024one},leaf,text width=30em]
        ]
    ]
    [Theory
            [Nguyen \textit{et al.} \cite{nguyen2024statistical}{,} Chen \textit{et al.} \cite{chen2022towards}{,} Li \textit{et al.} \cite{li2024theory1}{,} Li \textit{et al.} \cite{li2024theory2}{,} Chowdhury \textit{et al.} \cite{chowdhury2023patch}{,}Jiang \textit{et al.} \cite{jiang1999hierarchical}{,} Jiang \textit{et al.} \cite{jiang1999approximation}{,}\\ Zeevi \textit{et al.} \cite{zeevi1998error}{,} Nguyen \textit{et al.} \cite{nguyen2016universal}{,} Mendes \textit{et al.} \cite{mendes2012convergence}{,} Ho \textit{et al.} \cite{ho2022convergence}{,} Nguyen \textit{et al.} \cite{nguyen2023demystifying}{,} Nguyen \textit{et al.} \cite{nguyen2023statistical}{,} \\Nguyen \textit{et al.} \cite{nguyen2024towards}{,} Nguyen \textit{et al.} \cite{nguyen2024least}{,} Nguyen \textit{et al.} \cite{nguyen2023general}{,} Fung \textit{et al.} \cite{fung2025mixture},leaf,text width=37.5em]
    ]
    [Applications
        [CV
            [Image \\ Classification
                [V-MoE \cite{riquelme2021scaling}{,} Videau \textit{et al.} \cite{videau2024mixture}{,} Vimoe \cite{han2024vimoe}{,} Royer \textit{et al.} \cite{royer2023revisiting}{,} \\Clip-moe \cite{zhang2024clip}{,} SoftMoE \cite{puigcerver2023sparse}{,} DeepME \cite{he_deepme_2021}{,} Jiang \textit{et al.} \cite{jiang2024mixtral} {,}\\ Nguyen \textit{et al.} \cite{nguyen2024expert},leaf,text width=22.5em
                ]
            ]
            [Object \\  Detection
                [MoCaE \cite{oksuz2023mocae}{,} Wang \textit{et al.} \cite{wang2024object}{,} Damex \cite{jain2024damex}{,}  Feng \textit{et al.} \cite{feng2024pluralistic},leaf,text width=22.5em
                ]
            ]
            [Semantic \\ Segmentation
                [Pavlitskaya \textit{et al.} \cite{pavlitska2024towards}{,} Zhu \textit{et al.} \cite{zhu2024customize} {,} Pavlitskaya \textit{et al.} \cite{pavlitskaya_using_2020}{,} \\DeepMoE \cite{wang2020deep}{,} Swin2-MoSS \cite{rossi2025swin2},leaf,text width=22.5em
                ]
            ]
            [Image \\ Generation
                [RAPHAEL \cite{xue2023raphael}{,} MEGAN \cite{park2018megan}{,} MoA \cite{wang2024moa}{,} Text2human \cite{jiang2022text2human},leaf,text width=22.5em
                ]
            ]
        ]
        [NLP
            [NLU
                [GLaM \cite{du2022glam}{,} MoE-LPR \cite{zhou2024moe} {,} MoE-SLU \cite{cheng_moe-slu_2024}{,} MT-TaG \cite{gupta2022sparsely}{,} MoPE-BA \cite{wu2024mixture},leaf,text width=22.5em
                ]
            ]
            [NLG
                [Text Generation
                    [Chai \textit{et al.} \cite{10095401}{,} RetGen \cite{zhang2022retgen}{,} LogicMoE \cite{wu_enhancing_2024}{,} \\QMoE \cite{frantar2023qmoe},leaf,text width=15em
                    ]
                ]
                [Machine Translation
                    [Shazeer \textit{et al.} \cite{shazeer2017outrageously}{,} Gshard \cite{lepikhin2020gshard}{,} Team \textit{et al.} \cite{costa2022no}{,} \\NLLB TEAM \textit{et al.} \cite{nllb_team_scaling_2024}{,} Huang \textit{et al.} \cite{huang2023towards},leaf,text width=15em
                    ]
                ]
                [Multimodal Fusion
                    [LLaVA-MoLE \cite{chen2024llava}{,} LIMoE \cite{mustafa2022multimodal}{,} Sun \textit{et al.} \cite{sun2024hunyuan},leaf,text width=15em
                    ]
                ]
            ]
        ]
    ]
  ]
\end{forest}
\caption{The roadmap of Mixture of Experts (MoE) covered in this paper.}
\label{fig:taxonomy}
\end{figure*}
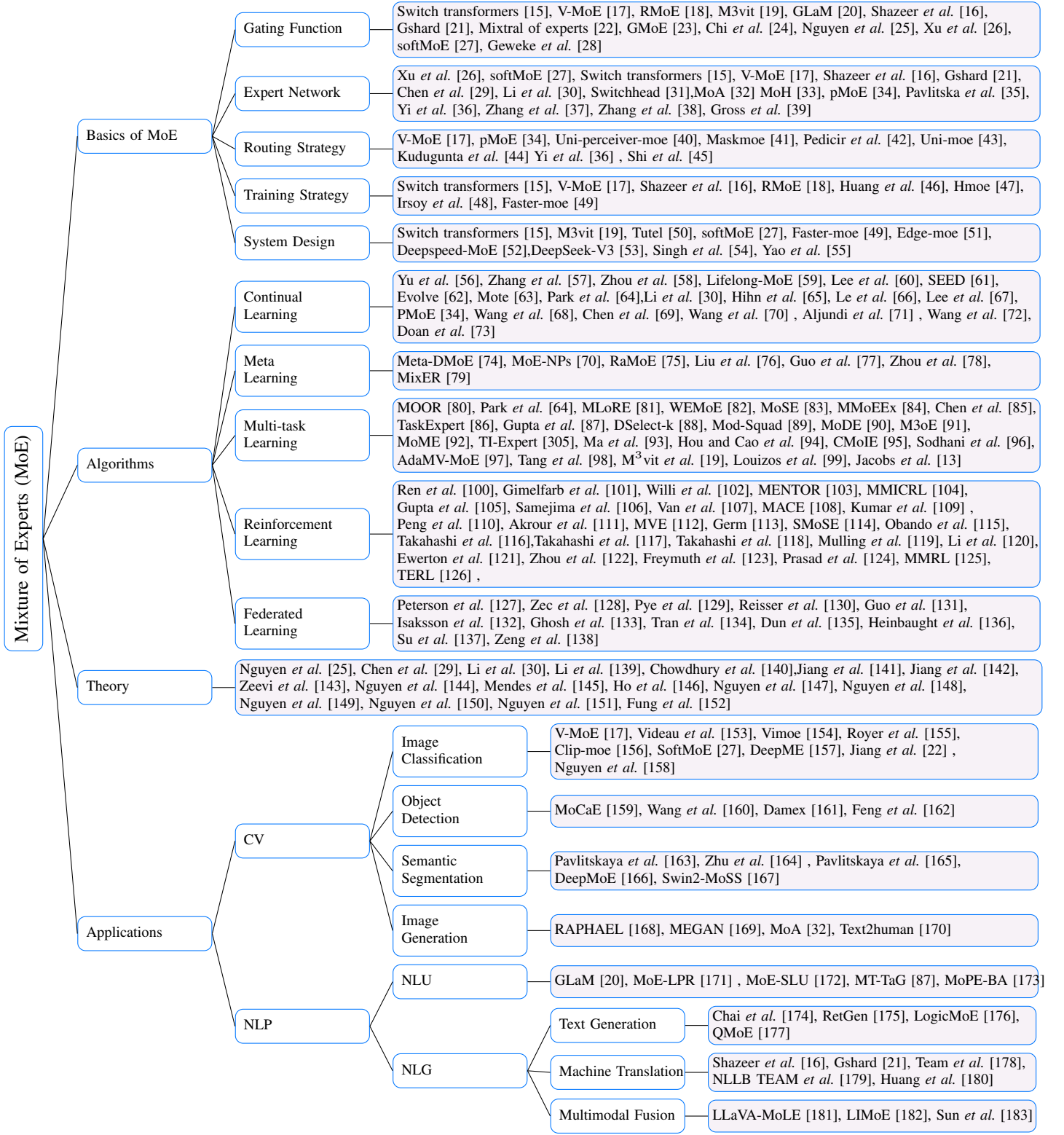

Notably, the MoE model architecture  has demonstrated its unique advantages and tremendous potentials especially in  large language models (LLMs) \cite{yang2024qwen2,wei2406skywork,gupta2024dbrx,lepikhin2020gshard,fedus2022switch,du2022glam,rajbhandari2022deepspeed,zoph2022st,zhu2024llama,qu2024llama,vavre2024llama,xue2024openmoe,lenz2025jamba,team2024jamba,lieber2024jamba,wu2024yuan}. For instance, the Switch Transformer \cite{fedus2022switch} achieves 7 times faster in pre-training speed compared to the T5-Base \cite{raffel2020exploring} model, and shows improved performance across all 101 languages in a multilingual setting. It has also successfully scaled the number of parameters to the trillion level, achieving a pre-training speed 4 times faster than the T5-XXL \cite{raffel2020exploring} model at this scale. GLaM \cite{du2022glam} has also expanded its parameters to the trillion level, enhancing the model's ability to utilize contextual information. Deepspeed-moe \cite{rajbhandari2022deepspeed} leverages MoE and model compression techniques to reduce the size of MoE models by 3.7 times. ST-MoE \cite{zoph2022st} has further investigated the stability and transferability of MoE models, proposing various methods such as Router z-loss to ensure training stability. Ultimately, on the SuperGLUE \cite{sarlin2020superglue} benchmark, the ST-MoE-32B model surpassed the previous state-of-the-art models. OpenMoE \cite{xue2024openmoe} has experimented with decoder-only MoE and provided an in-depth discussion on the routing mechanisms of MoE, making significant contributions to the open-source community's understanding of MoE architectures. Mixtral 8×7B \cite{jiang2024mixtral}, despite processing each token with only 13 billion active parameters, is able to access a total of 47 billion parameters thanks to its distinctive architecture. This allows Mixtral 8x7B to achieve higher parameter efficiency and effectively control computational costs. The impressive performance of the DeepSeek series \cite{dai2024deepseekmoe,deepseek-ai_deepseek-v2_2024,liu2024deepseek,deepseek-ai_deepseek-r1_2025} has also garnered significant attention. These models leverage MoE architectures to achieve state-of-the-art results on a variety of benchmarks while maintaining manageable computational requirements.

Beyond efficiency, MoE models also offer opportunities to improve model interpretability \cite{pmlr-v161-gimelfarb21a,RL_17,pavlitska_sparsely-gated_2023,fung_class_2019}. By learning intrinsic allocation mechanisms, researchers can gain insights into how different experts specialize in handling specific types of data or tasks. This interpretability not only enhances our understanding of the model behaviors but also opens new avenues for designing more robust and transparent AI systems.

Given the rapidly growing research interests in MoE and the huge potentials of MoE in various application domains, there is an urgent need of a comprehensive survey for summarizing and disseminating the recent advances of MoE, in order to elicit escalating attentions and inspire further research ideas in this field. Nevertheless, the early surveys of MoE \cite{yuksel2012twenty,masoudnia2014mixture} were published ten years ago and have not incorporated the new development in the area. There is only one new survey \cite{cai2024survey} that emerged last year, which however 1) emphasized more on the basic designs of MoE  and 2) did not provide a broad discussion on the use of MoEs in important machine learning paradigms and application fields such as computer vision. The theory development in MoE has also not been covered. 

And although some other new concurrent surveys  have been uploaded online very recently during the preparation of our paper, these surveys still either focus on a special aspect of MoE, such as applications in big data \cite{gan2025mixture}, inference optimization \cite{liu2024survey} and LLMs \cite{vats2024evolution},  or cover multiple topics very briefly \cite{dimitrisurvey}. To the best of our knowledge, our paper is the first survey that comprehensively summarizes the latest advancement in MoE, which includes four key components: basic design strategies of MoE, MoE-based algorithm designs in mainstream machine learning directions,  theory towards understanding MoE in various scenarios, and an in-depth review of applications of MoE in both computer vision and natural language processing.

This paper will be organized as follows (as illustrated in \Cref{fig:taxonomy}). In \Cref{sec: sec_2}, we will provide a comprehensive introduction to the basic designs of MoE, ranging from basic architecture design to training strategies to system optimization strategies. In \Cref{sec:sec_3}, we will introduce the recent MoE-based algorithm design in four important machine learning paradigms, including continual learning, meta-learning, multi-task learning, reinforcement learning, and federated learning, which seeks to provide readers a general idea about how MoE can be leveraged to improve the algorithm design in these domains. In \Cref{sec:sec_4}, we will summarize the efforts towards building the theoretical understanding of MoE, which covers a variety of aspects in MoE, such as different gating functions, different expert models, and different learning scenarios. In \Cref{sec:sec_5}, we will present the recent applications of MoE in two major domains, i.e., computer vision and natural language processing, where we look into multiple important subproblems in each domain. Future research directions will be discussed in \Cref{sec:sec_6}, followed by the conclusion in \Cref{sec:sec_7}.

\section{Basic Designs of MoE}\label{sec: sec_2}

\begin{figure}[htbp]
    \centering
    \includegraphics[width=0.5\textwidth]{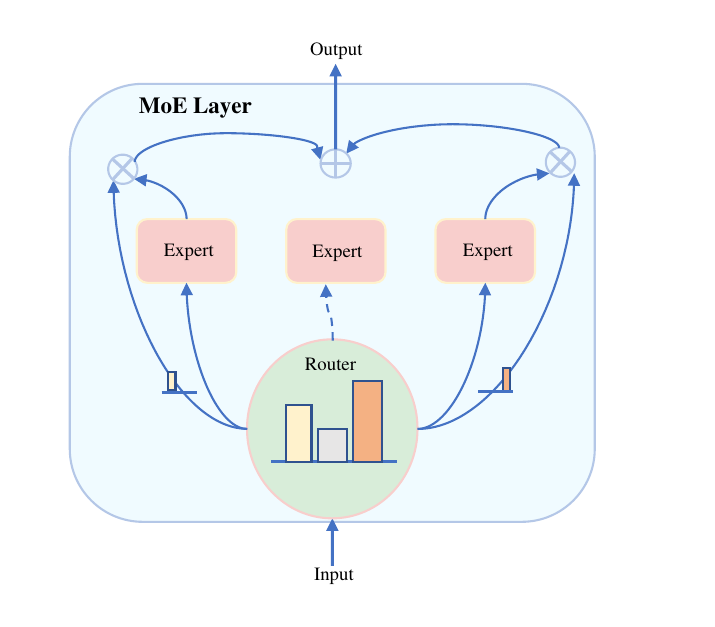}
    \caption{The simple schematic of a standard MoE architecture.}
    \label{fig:arch}
\end{figure}

In this section, we will focus on the basic framework and design considerations in MoE to help readers understand the general workflow of MoE, as shown in \Cref{fig:arch}. More specifically, we will
first introduce
the key components in the design of MoE, including the gating function, the expert networks, and the routing mechanism.
 The first two components essentially form a basic framework of MoE, 
 whereas the last component characterizes how the gating function handles the input data.
Next, we will introduce the  strategies, i.e., loss function and pipeline design, so as to guarantee proper training of MoEs. 
In the end, system designs will be discussed to optimize the system efficiency of MoE from multiple perspectives.

\subsection{Gating Function}
The gating function serves as the mathematical implementation of a router, determining how input data is allocated to designated experts. Clearly, the effectiveness of MoE models is intrinsically tied to this allocation strategy, making the design of the gating function a critical consideration. Below, we provide a detailed discussion on the principles and considerations for designing an effective gating function.

When selecting a gating function, the following criteria should be prioritized: (1) The function must accurately discern the characteristics of both input data and experts. This enables the assignment of similar data to the same expert or group of experts, ensuring that each expert receives a sufficiently large and coherent training set. Such specialization allows experts to develop expertise in specific  knowledge domains. (2) The input data should be distributed as evenly as possible among the predefined  experts. An uneven distribution can lead to model collapse, which reduces the efficiency of the MoE framework by underutilizing experts and impairs performance due to insufficient separation of conflicting knowledge in the training data. 

\textbf{1) Linear Gating.} Based on these considerations, most existing MoE models \cite{fedus2022switch,riquelme2021scaling,wu2022residual,fan2022m3vit,du2022glam} use a linear function with softmax as their gating functions due to its simplicity and effectiveness, which is also referred as softmax gating:
\begin{equation}\label{eq:linear}
G(x)_i = \text{softmax}(\text{TopK}(g(x) + \mathcal{R}_{\text{noise}}, k))_i,
\end{equation}
\begin{equation}
    \text{TopK}(v) = 
    \begin{cases} 
      v_i & \text{if } v_i \text{ is one of the top } k \text{ elements,} \\
      -\infty & \text{otherwise.}
    \end{cases}
    \label{eq:top_k}
\end{equation}

Here $G$ represents the gating function, the $\text{TopK}$ function retains the $k$ highest scoring inputs out of $N$ experts and sets the rest to negative infinity, $g$ represents the gating value calculated with a linear function prior to the softmax operation, $x$ denotes the input, $\mathcal{R}_{\text{noise}}$ represents the noise to encourage expert exploration, $w$ denotes the model parameters, and $k$ is a hyperparameter that can either be learned or manually set. 

The order of the TopK and softmax operations in the above expression is flexible and can be designed based on specific requirements. One approach is to perform the TopK operation before softmax \cite{shazeer2017outrageously,jiang2024mixtral,du2022glam}. This method quickly filters out the most relevant experts, eliminating the need to compute softmax for all experts and thereby reducing computational overhead. However, the scores obtained after TopK may not conform to a probability distribution, necessitating an additional normalization step. Moreover, the TopK operation is inherently ``hard," potentially excluding experts with low scores that could still contribute meaningfully.
Alternatively, the TopK function can be applied after softmax \cite{lepikhin2020gshard,riquelme2021scaling,fedus2022switch,wu2022residual}. In this case, the softmax function first normalizes the scores into a probability distribution, providing statistically meaningful activation weights for each expert. This approach offers more clear guidance on determining the value of $k$ in the TopK operation. However, it requires computing softmax for all experts, resulting in higher computational costs.

When the MoE is incorporated into existing models, usually a Transformer \cite{vaswani2017attention}, we typically replace the Feed-Forward Network (FFN) layer within a transformer block by using an MoE layer. There are several reasons \cite{shazeer2017outrageously} behind this design choice: First, the computational cost of the FFN layer typically constitutes a significant portion of the total computational cost in a Transformer, especially in deep models. By replacing the FFN layer with an MoE layer, the computational cost can be significantly reduced while maintaining the model's expressive power. Additionally, replacing the Self-Attention layer would disrupt the core mechanism of the Transformer, which is not aligned with our objectives.

\textbf{2) Non-linear Gating.} The gating functions can also be non-linear. Specifically,
a gating function design based on cosine distance is proposed in GMoE~\cite{li2022sparse} for domain generalization tasks, which is shown as follows:
\begin{equation}
    G(x)= \text{TopK}\left(\text{softmax}\left(\frac{E^T W_{linear}x}{\tau \|W_{linear}x\|\|E\|}\right)\right)\qquad 
\end{equation}
where $x \in \mathbb{R}^{d_e}$ is the input of MoE module, $W_{linear} \in \mathbb{R}^{d_e}$ is a learnable linear transformation for $x$, responsible for projecting $x$ into a hypersphere space. The number of expert networks is $N$, and $E \in \mathbb{R}^{d_e \times N}$ represents the features of the $N$ expert networks. In the gating function, the input $x$ is projected into a new space and compared with the expert embeddings $E$ using cosine similarity to determine which experts each input $x$ should be assigned to. In this way, the expert embeddings $E$ can capture the feature representations of different experts and are gradually optimized during training to better adapt to the task requirements. Additinally, the temperature parameter $\tau$ controls the sharpness of the gating distribution. A smaller $\tau$ makes the distribution sharper, favoring the selection of a few experts, while a larger $\tau$ makes the distribution smoother, allowing more experts to participate.

Comprehensive theoretical and experimental evidence has been provided in \cite{li2022sparse} to demonstrate the superiority of using cosine routers over linear routers in domain generalization tasks. Specifically, the cosine router excels at handling cross-domain data, capturing visual attributes, and enhancing the model's generalization capability and efficiency. There are also more studies that have employed cosine routers for better performance \cite{chi2022representation,nguyen2024statistical}, which encourages us to explore more on the design of gating functions beyond the cosine routers.

In addition to the cosine gating function, there are many other designs of nonlinear functions, such as the gating function based on exponential Family of Distributions \cite{xu1994alternative} as shown in \Cref{eq:pGate}, 
\begin{equation}\label{eq:pGate}
    G_j(x, \nu) = \frac{\beta_j D(x | \nu_j)}{\sum_i \beta_i D(x | \nu_i)},
\end{equation}
Here $\beta_j$ is the prior probability of the $j$th expert network, satisfying $\sum_j \beta_j = 1$ and $\beta_j \geq 0$. $D(x | \nu_j)$ is the conditional probability density function of input $x$ under the $j$th expert network, belonging to the exponential family distribution.
This approach analytically updates the gating function during iterations, and overcomes the nonlinear relationships introduced by softmax and the additional iterative optimization steps that follow (such as IRLS, Iteratively Reweighted Least Squares). The most common example is the Gaussian distribution.

 There is also a special gating function in Soft MoE \cite{puigcerver2023sparse} that no longer employs a discrete token allocation mechanism. Specifically, Soft MoE calculates the weights between each token and each expert, then uses these weights for weighted averaging to generate input slots. Consequently, the slots processed by each expert are a weighted combination of all tokens, rather than individual tokens. This approach avoids potential token dropping issues due to load imbalances and optimization difficulties caused by discrete operations such as TopK.

Gating functions based on the Student-t distribution \cite{ingrassia2012local} are particularly suitable for data groups exhibiting long-tail phenomena or outliers, because they offer more robust fitting capabilities. \cite{geweke2007smoothly} utilizes a multinomial probit model \cite{geweke1994alternative,keane1992note} to determine the expert model that each observation belongs to.


\begin{figure}[htbp]
    \centering
    \includegraphics[width=0.5\textwidth]{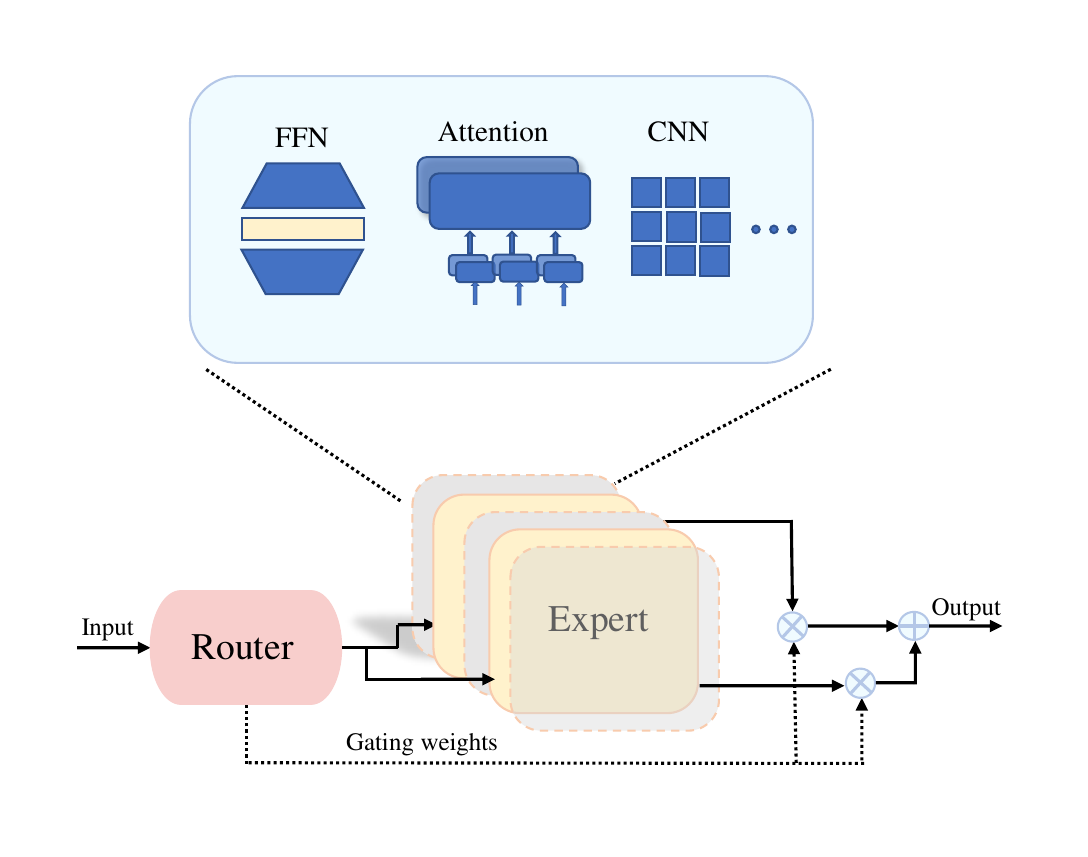}
    \caption{Various expert networks for the MoE layer schematic. }

    \label{fig:basics}
\end{figure}

\subsection{Expert Networks}

The expert network is a core component of the MoE architecture. By dynamically selecting the most suitable expert network through the gating function, MoE efficiently allocates input data, enabling different experts to specialize in distinct knowledge domains. This design enhances the model's overall performance, generalization ability, and computational efficiency, particularly when handling complex and diverse tasks.

In principle, each expert network in MoE can function as an independent network model, similar to a single network model (e.g., \cite{chen2022towards,li2024theory1}). However, in practice, to ensure efficiency and scalability, expert networks are often integrated into a single network model, with specific layers replaced by MoE layers \cite{shazeer2017outrageously,fedus2022switch,lepikhin2020gshard}. Notably, MoE layers can be introduced at specific levels of existing neural network models without altering the overall structure, demonstrating high flexibility and broad applicability. Currently, the most widely used MoE layers include the following types (\Cref{fig:basics}):

1) \textbf{Replace the FFN layer in Transformer with an MoE layer}. The MoE layer can be characterized as follows:

\begin{equation}
    \label{eq:moe_ffn}
    \text{MoE}(x) = \sum_{i \in I\!D} w_i M_i(x)
\end{equation}

where $I\!D$ represents the set of indices of experts selected by the gating function, $w_{i}$ is the weight assigned to the ith expert, and $M_{i}(x)$ defines a two-layer FFN expert model.
 One key reason for introducing the MoE mechanism into the FFN layer is due to the higher sparsity and  domain specificity of the FFN layer in Transformer. It has been shown \cite{riquelme2021scaling} that the FFN layer exhibits higher sparsity compared to self-attention layers,  mainly because not all units in the fully-connected FFN layer will be activated simultaneously in order to prevent overfitting. This will make the FNN layer more suitable to incorporate sparse activation mechanisms. Furthermore, \cite{riquelme2021scaling} reveals the Emergent Modularity phenomenon in pre-trained Transformer models, indicating a significant association between neuron activations and specific tasks. This further motivates  the idea of leveraging the MoE structure to reflect the modular nature of pre-trained Transformers. Therefore, due to its inherent sparsity and task relevance, the FFN layer in Transformer becomes an ideal choice for introducing the MoE mechanism.

2) \textbf{Apply MoE to the attention module in Transformer}. 
This will lead to a special type of MoE, namely mixture-of-attention (MoA) \cite{wang2024moa}. Specifically,
MoA consists of a set of attention heads with different parameters, where each attention head can be regarded as an expert ${f_1,...,f_N}$. For a given input, the gating function selects the TopK attention heads based on the importance scores of the input, and  the final output is generated as a weighted sum of only these selected heads' outputs. This design allows the model to focus on the most relevant attention heads, enhancing performance without significantly increasing computational costs, and is highly scalable to support larger parameter sizes. 

Compared to the standard multi-head attention mechanism \cite{vaswani2017attention}, the key advantage of MoA lies in its use of a gating network to dynamically select the  most relevant attention heads for each input token and compute a weighted sum of their outputs. This approach eliminates the requirement for all attention heads to process every input, enabling MoA to achieve superior model performance with reduced computational overhead. Consequently, MoA exhibits enhanced scalability, and particularly this model can attain higher BLEU scores \cite{papineni2002bleu} than Transformer  with 213 million parameters  while requiring significantly less computation. 

Inspired by MoA, numerous studies have been proposed to improve upon this design. SwitchHead \cite{csordas2023switchhead} attempts to independently introduce the MoE mechanism for each head's key, query, value, and output projection, and employs a non-competitive selection activation function. This will therefore reduce the number of attention matrices that need to be computed, significantly lowering computational and memory requirements. MoH \cite{jin2024moh} enhances the standard MoE method by sharing heads and a two-stage routing strategy, and can further fine-tune pre-trained multi-head attention models into MoH models, enhancing their applicability. In addition, this work integrates the proposed MoH framework with various model frameworks including the encoder-decoder architecture, such as ViT \cite{dosovitskiy2020image}, DiT \cite{peebles2023scalable}, and decoder-only LLMs \cite{mann2020language}, which are applied in various tasks and demonstrate great potentials of MoH.

3) \textbf{Apply MoE to CNN layers.} Integrating CNN into the MoE architecture can fully leverage CNN's strengths in local feature extraction, achieving more refined task allocation and improved computational efficiency. As a result, many studies have attempted to effectively combine MoE with CNN. For example, in \cite{chowdhury2023patch}, a simple pMoE structure is adopted, where the MoE layer consists of a gating kernel (a trainable weight network) and multiple expert networks composed of two-layer CNN networks. 

Image data possesses high local correlation and spatial hierarchical structure, which aligns well with the design philosophy of CNNs. As a result, Convolutional Mixture of Experts (CMoE) is often applied to computer vision tasks. For instance, \cite{zhang_learning_2019} utilizes the MoE framework to decompose fine-grained classification problems into subspace problems, allowing each expert CNN network to focus on different sub-problems and improving classification performance. Building on \cite{zhang_learning_2019}, \cite{zhang_enhancing_2022} proposes an attention-based MoE framework to enhance the model's generalization ability and classification performance under limited data conditions. \cite{gross2017hard} addresses the issue of parallel training of independent expert networks in MoE models, alleviating the need for multiple GPUs in large-scale supervised visual tasks. Meanwhile, \cite{pavlitska_sparsely-gated_2023} focuses on exploring how to enhance the interpretability of CNNs while maintaining model performance in computer vision tasks. Additionally, a small number of works have explored the potential of CMoE in other domains, such as \cite{yi_mixture_2019}.

\subsection{Routing Strategy}

A critical consideration in the design of MoE is determining the level at which experts should specialize in knowledge. For instance, in computer vision, certain tasks may emphasize more on the global features, while others demand precise local perception. Consequently, selecting an appropriate routing level—defined by the frequency of routing decisions—is essential to ensure the MoE model effectively meets the requirements of specific tasks. The following discussion extends the classification strategy proposed in Uni-Perceiver-MoE \cite{zhu2022uni} and outlines several common routing strategies: token-level routing, modality-level routing, task-level routing, and other-level routing.

\textbf{1) Token-Level Routing Strategy} determines routing decisions based on token representations, making it the most classic routing approach. Tokens also come in various types, among which the most common ones we encounter are text tokens and image tokens (patch tokens). 

In terms of text tokens,
to mitigate underfitting and preserve representation diversity, MaskMoE~\cite{su2024maskmoe} introduces a novel token-level routing strategy. This approach generates a mask vector for each token in the vocabulary and adjusts the number of visible experts for tokens of varying frequencies before training. For infrequent tokens, MaskMoE restricts routing to a single expert via routing masks, ensuring consistent training for these tokens. For frequent tokens, it permits multiple visible experts to capture contextual nuances. Additionally, \cite{pedicir2024novel} proposes a token-level recurring routing strategy that employs recurrent neural networks to dynamically adjust expert selection based on input sequence context. This strategy improves the model's accuracy in processing complex linguistic structures, optimizes computational resource utilization, and effectively addresses long-range dependencies. Consequently, it achieves enhanced efficiency and performance across diverse natural language processing tasks.

The usage of patch tokens, as discussed in \cite{chowdhury2023patch}, is often built upon the ViT architecture. This strategy divides an image into multiple patches and dynamically routes each patch to the most suitable expert for processing. By adaptively selecting experts based on the local features of the image, it optimizes computational resource utilization and enhances the model's performance in handling complex visual tasks. Additionally, V-MoE~\cite{riquelme2021scaling} introduces a Batch Priority Routing (BPR) strategy, which 1) calculates the importance of each image patch, 2) sorts them in descending order of importance, and 3) assigns experts to patches accordingly. Experimental results demonstrate that BPR can maintain performance comparable to a dense model while processing only 15\%-30\% of the image patches, showcasing significant advantages by standard Routing.

 Apart from these two common types, there are also audio tokens \cite{yi_mixture_2019}, time series token  \cite{shi2024time}, etc., all of which can be freely combined with MoE to help the model achieve better learning performance.

\textbf{2) Modality-Level Routing Strategy} routes tokens based on their modality, emulating the practice of employing specialized encoders for distinct modalities to minimize interference between them. The Uni-MoE model~\cite{li2024uni} is designed to effectively process and understand multi-modal data through a structured approach. First, it introduces connectors trained using cross-modal data to achieve data alignment, mapping diverse modalities into a unified language representation space. Simultaneously, it trains modality-specific expert systems on cross-modal instruction datasets, so as to enhance each expert's specialized capabilities and activate their task-specific preferences. Subsequently, the model utilizes LoRA \cite{hu2022lora} for fine-tuning, optimizing performance with mixed multi-modal instruction data while maintaining computational efficiency. Following instruction tuning, Uni-MoE is evaluated on a comprehensive and diverse multi-modal dataset. The results indicate that the model significantly reduces performance discrepancies when processing mixed data, improves collaboration efficiency among multiple experts, and enhances generalization capabilities, highlighting the potential of the MoE architecture in developing a unified multi-modal large language model.

\textbf{3) Task-Level Routing Strategy} determines routing based on task IDs, ensuring that different tasks are routed to distinct experts, thereby effectively minimizing interference between tasks. This approach is particularly advantageous for multi-task learning. For instance, \cite{kudugunta2021beyond} explores multilingual machine translation tasks, where task boundaries are defined by target languages, language pairs, or unique task IDs. And the experts are selected at the task level according to these boundaries. Compared to token-level routing strategies, this approach only requires loading a subset of experts relevant to the current task during inference, rather than loading all experts of the entire model. This results in reduced communication costs between devices and lower memory usage.

\textbf{4) Other-Level Routing Strategies} encompass a variety of approaches, such as context-level and attribute-level routing strategies~\cite{zhu2022uni}. The context-level routing strategy employs global pooling to provide global context information to the router, enabling it to generate reliable routing strategies based on the contextual information of the current token representation. In contrast, attribute-level routing incorporates an 8-dimensional binary embedding, which includes attributes such as the modalities of the current task and token (indices 0-5), the causation type of the model (index 6), and the token source (index 7). These rich information enables models to handle more complex and different tasks. 

\subsection{Training Strategies}
Due to the sparse activations inherent in MoE-based architectures, directly applying the functions and procedures used for training dense models to MoE-based sparse models is not straightforward. To improve training effectiveness, achieve faster convergence, and enhance performance on target tasks, numerous studies like \cite{fedus2022switch,riquelme2021scaling,shazeer2017outrageously,wu2022residual,huang2024harder} have explored the methods for training these sparse models. In the following section, we summarize and discuss some representative techniques.

\textbf{1) Auxiliary Loss Function Design.} In addition to standard training losses, auxiliary loss functions are often introduced during MoE training to enhance learning performance, improve model efficiency, and stabilize the training process. One type of  auxiliary losses widely used in practice is the load balancing loss, which aims to balance the allocation of inputs across all experts to fully utilize the capacity of the MoE.

A well-known challenge in MoE training is the risk of model collapse, where a small subset of experts receives the majority of inputs, while others receive almost none, leading to an imbalance in expert utilization. To mitigate this issue, \cite{shazeer2017outrageously} proposes Importance and Load loss functions, which are defined as follows:

\begin{equation}
\text{Load}(X)_i = \sum_{x \in X} P(x, i) \\
\end{equation}
\begin{equation}
\text{Importance}(X)_i = \sum_{x \in X} G(x)_i \\ 
\end{equation}
\begin{equation}
L_{\text{load}}(X) = w_{\text{load}} \cdot \left( \text{CoV}(\text{Load}(X))\right)^2 \\
\end{equation}
\begin{equation}
L_{\text{importance}}(X) = w_{\text{importance}} \cdot \left( \text{CoV}(\text{Importance}(X)) \right)^2 \\
\end{equation}

where \(w_{importance}\) and \(w_{load}\) are hyperparameters controlling the influence of expert importance balance and load balancing, respectively, and \(P(x,i)\) denotes the probability that \(G(x)_i\) is non-zero given new random noise selections. Here, \(\text{CoV}\) stands for the coefficient of variation. These two loss functions are incorporated into the overall model loss function to promote balanced usage and load distribution among experts, thereby enhancing model performance and stability.

The Switch Transformer \cite{fedus2022switch} (as shown in \Cref{eq:simp}) simplifies the  approach above based on \cite{shazeer2017outrageously}, trading a slight reduction in accuracy for improved efficiency: 
\begin{equation}
    f_i = \frac{1}{T} \sum_{x \in \mathcal{B}} \mathbbm{1}\left\{ \operatorname{argmax} G(x) = i \right\}\, 
\end{equation}
\begin{equation}
    Q_i = \frac{1}{T} \sum_{x \in \mathcal{B}} G(x)_{i},
\end{equation}
\begin{equation}\label{eq:simp}
    \text{loss} = \alpha \cdot N \cdot \sum_{i=1}^{N} f_i \cdot Q_i.
\end{equation}

Here $ T $ is the total number of tokens in the batch, $ \mathcal{B} $ is the set of tokens in the batch, $ f_i $ is the proportion of tokens assigned to expert $ i $, $ Q_i $ is the routing probability ratio assigned to expert $ i $, and $ \alpha $ is a hyperparameter used to control the weight of the auxiliary loss.
Ideally, all experts should receive tasks evenly, with each processing an equal amount of input data, such that the expected values of \(f\) and \(Q\) are both \(1/N\). When the distribution is uniform, this auxiliary loss function reaches its minimum. Expert capacity \cite{shazeer2017outrageously} refers to the maximum number of data points each expert can process. Insufficient expert capacity may result in skipped input data during training, leading to under-trained experts; conversely, increasing capacity reduces dropped tokens but may also waste computational and memory resources. V-MoE \cite{riquelme2021scaling} normalizes \(L_{Importance}\) and \(L_{Load}\) when applying the auxiliary loss function to visual tasks, achieving a more stable training process.

\textbf{2) Expert Selection.}
It is clear that the performance and efficiency of MoE models heavily depend on the expert selection for data inputs. While the gating function primarily determines expert selection, additional design choices can further enhance MoE models. The TopK strategy, which selects the best $K$ experts among all experts to process the data, 
significantly increases model capacity without proportionally escalating computational costs. For example,  \cite{shazeer2017outrageously} has shown that the MoE model can scale to over 1000 times of the size of traditional neural networks while maintaining computational efficiency. 
On the other hand, the TopK strategy can be removed in order to further enhance the efficiency. For instance, V-MoE \cite{riquelme2021scaling} replaces TopK with a Top-1 strategy to simplify computation and reduce resource wastage. 
Moreover, randomness is usually introduced for expert selection inside the TopK function (as shown in \Cref{eq:linear}), 
 by adding Gaussian noise to expert activations during the selection process \cite{shazeer2017outrageously}. This mechanism encourages 
 expert exploration with random selections and 
prevents over-reliance on specific experts, which mitigates load imbalance and optimizes resource utilization.

Dynamic expert selection mechanisms have also been explored in the literature. Notably, \cite{huang2024harder} proposes a Top-P routing strategy, which determines the number of activated experts at each time by setting a probability threshold \(P\). For each input data, the gating network calculates a score for each expert, corresponding to a probability distribution of an expert being selected, and  the experts are sorted in a descending order of their scores. The strategy then progressively accumulates the scores of the experts until the cumulative value exceeds \(P\), activating all accumulated experts to process the token. This  allows the model to adaptively adjust the number of selected experts based on input complexity, enabling flexible and efficient resource allocation.
To reduce overfitting and improving generalization,  \cite{irsoy2021dropout} proposes a dropout regularization method tailored for the tree structure of Hierarchical Mixture of Experts (HMoE) \cite{wang2024hmoe}, which  randomly drops expert networks of different branches according to the hierarchical structure. Experimental results demonstrate that this method enhances model performance, providing smoother fitting outcomes, better generalization, and lower validation error rates compared to models without dropout.

\textbf{3) Pipeline Design.}
Given  the sparsity nature and need of dynamic data allocation in MoE architectures,  a well-designed pipeline plays a crucial role in the training of MoE models, in order to
maximize the utilization of computational resources, reduce time and costs, and ensure stable convergence of the model. In general, the pipeline design for MoE seeks to optimize resource allocation and efficiently distribute data among experts. 

As mentioned before, BPR \cite{riquelme2021scaling} achieves sparsity reuse by prioritizing important samples for processing. The advantages of BPR lie in its efficient resource utilization and reduction of unnecessary computations. RMoE~\cite{wu2022residual} designs a pipeline particularly for downstream tasks such as segmentation and detection. More specifically, RMoE decomposes the weights of MoE experts into core and residual weights, where the core weights inherit knowledge from pre-trained non-MoE models and only the residual weights will be finetuned. This will significantly shorten the training time and reduce the training costs. During inference, both sets of weights will be combined to ensure the good performance of the MoE model.


\subsection{System Design}
The MoE architecture is gaining popularity in the edge deployment and capacity scaling of large models \cite{fedus2022switch,sarkar2023edge,rajbhandari2022deepspeed}, offering significant flexibility and scalability. However,
its nature of inherent non-uniform workload distribution and dynamic expert selection imposes strict requirements on the system's resource allocation, communication efficiency, and overall performance optimization. This will introduce new system challenges such as increased synchronization overhead, hampering the efficiency and stability of model execution. In this section, we outline some common methods currently available to enhance the system designs for MoE from three dimensions: computation, communication, and memory.

\textbf{1) Computation.}
In the application of MoE models, due to their inherent dynamic characteristics, there may be significant disparities in the number of tasks processed by different experts, making the issue of non-uniform workload distribution particularly pronounced. Particularly, in large-scale distributed system environments, standard MoE architectures require synchronization among expert operations \cite{he2022fastermoe}, i.e., all experts must complete their current tasks before moving on to the next computation step. However, some experts may finish work earlier due to smaller assigned task volumes, leading to idle timeslots and wasted computational resources. Moreover, as the MoE model capacity expands, computational complexity significantly increases \cite{fedus2022switch}, intensifying the demand for better computation strategy design.

Various attempts have been made in the literature to mitigate this problem. The paper \cite{fedus2022switch} proposes and discusses several parallel strategies, including data parallelism, expert parallelism, and model parallelism, but different parallel methods have varying requirements for the distribution of data and model parameters. To enable the system to quickly adapt to dynamic load changes, \cite{hwang2023tutel} designs a unified distribution layout to manage the model's parameters and input data. This can accommodate the distribution requirements of various parallel strategies, achieving zero-cost switching of parallel strategies. Based on this distribution, \cite{hwang2023tutel} also further develops multiple optimization techniques, such as the ALL-to-ALL optimization algorithm mentioned later.
Moreover, Soft MoE~\cite{puigcerver2023sparse} introduces an improved scheduling algorithm that uses gradient-based routing selection methods, making the computation process smoother and more efficient. This approach reduces unnecessary synchronization wait times, lowers overall synchronization overhead, and boosts system throughput and efficiency. DeepSeek-V3 \cite{liu2024deepseek} introduces several enhancements to the communication process. Firstly, it concurrently processes two micro-batches with comparable computational loads, overlapping the attention mechanism and MoE component of one micro-batch with the dispatch and combine operations of another. This technique not only boosts throughput but also masks the overhead associated with all-to-all and tensor parallel communications, thereby substantially diminishing reliance on communication bandwidth. Furthermore, an FP8 format is employed for low-precision communication; activations are quantized to FP8 prior to up-projection in MoE using integral power-of-two scaling factors to mitigate quantization errors, a method similarly applied to activation gradients before down-projection in MoE. Lastly, to alleviate all-to-all communication bottlenecks, particularly among GPUs, a more efficient processing strategy has been devised. The specific implementation is available as open-source on GitHub at \url{https://github.com/deepseek-ai/DeepEP}.

\textbf{2) Communication.}
In the training of MoE models, input data must be distributed to the experts via an All-to-All communication pattern, which usually results in a significant demand on the communication bandwidth. On the other hand, due to the dynamic nature of expert selection, 
the amount of communication in the system can substantially fluctuate, such that the communication demands between iterations can also be very different. Clearly, this communication pattern is difficult to predict and optimize, thereby increasing the risk of network congestion.

Considering that only a portion of MoE models are activated at any given time, on-demand loading strategies \cite{fan2022m3vit} can be adopted when scheduling the model, i.e.,  weights are loaded into the on-chip memory only when the corresponding expert is selected. This method reduces frequent access to external DRAM, effectively alleviating the pressure on memory bandwidth. Moreover, different experts within the MoE can be assigned to distinct compute units or cores, achieving task-level parallel processing \cite{fedus2022switch}. This strategy not only increases  the system throughput but also fully exploits the parallelism provided by Multi-Chip Modules (MCMs). In addition to expert-level parallelism, other studies have proposed various parallelization strategies such as data parallelism, model parallelism or tensor parallelism \cite{fedus2022switch}, allowing more flexible design and the use of different parallel methods according to task characteristics \cite{hwang2023tutel,singh2023hybrid,yao2024exploiting}. To further optimize performance, systems such as TUTEL \cite{hwang2023tutel} and FasterMOE \cite{he2022fastermoe} simultaneously run the communication operations with other computational tasks to hide communication latency and reduce the impact of communication on overall runtime. Moreover, 
hardware features have also been utilized to motivate various communication patterns to reduce the need of bandwidth, e.g., 
EdgeMOE \cite{sarkar2023edge} and M3ViT \cite{fan2022m3vit}.

\textbf{3) Memory.}
The MoE model typically contains a large number of parameters due to the usage of multiple experts. This will need a large amount of memory, which sometimes may even exceed the storage capacity of a single device, thereby limiting the scalability of the MoE model. Moreover, due to the randomness and dynamics in expert selection, the memory access patterns of MoE models can become extremely complex.

To address these issues, Switch-Transformer \cite{fedus2022switch} proposes a parameter migration strategy that leverage the memory space of multiple devices in a collective manner, to store a large volume of parameters through parallel communication mechanisms. This method effectively distributes the pressure of parameter storage, enhancing the system's overall storage capacity and computational efficiency. DeepSpeed-MoE \cite{rajbhandari2022deepspeed} adopts a hierarchical storage management scheme, integrating local high-speed caches with remote slower storage to construct a multi-level storage architecture. This design ensures that frequently used parameters can be accessed quickly while also accommodating less frequently used but necessary parameters, thus achieving more efficient memory utilization without compromising performance.

\section{Algorithms} \label{sec:sec_3}

Due to the effectiveness and the strength in handling diverse knowledge in data, there have been rapidly growing interests in applying the MoE architecture into various machine learning paradigms, leading to improved algorithm designs. To showcase this and the great potentials of MoE in solving more complex learning scenarios, we will delve into the MoE-based algorithm designs for four widely investigated domains, i.e., continual learning, meta-learning, multi-task learning,  reinforcement learning, and federated learning. 

\subsection{Continual Learning}

Continual learning \cite{lin2022trgp} seeks to build an agent that can continuously learn a sequence of different tasks, corresponding to a non-stationary environment with data distribution shifts, to mimic the extraordinary lifelong learning capability of human beings. In particular, in continual learning a model is continuously adapted based on the new data without access to previous data. This will lead to a key challenge in continual learning, the so-called ``Catastrophic Forgetting" \cite{lin2022beyond}, which refers to the phenomenon that neural networks can easily forget the  previously acquired knowledge when learning the new task.  Seeing that the MoE framework dynamically selects the most appropriate sub-model for each input, enabling the model to handle diverse tasks without a substantial increase in computational overhead, multiple efforts have been made in the literature to leverage MoE to facilitate better continual  learning algorithm designs.


The idea of applying MoE in continual learning is first explored \cite{lee2020neural} which builds a model with a set of experts for task-free continual learning \cite{aljundi2019task}, i.e., a more general continual learning setup where the explicit task definition and task boundary information are unavailable. In particular, by following the expansion-based continual learning approaches, a Continual Neural Dirichlet Process Mixture (CN-DPM) model is introduced, where each expert model in CN-DPM handles a subset of data and the number of expert models dynamically expands based on the Bayesian nonparametric framework. 
However, the gating mechanism to determine the expert selection for current input data in \cite{lee2020neural} relies on a generative model for each expert. \cite{hihn2021mixture} improves over the design by showing that a gating mechanism based only on the input data is feasible. In particular, \cite{hihn2021mixture} considers a hierarchical variational continual learning (HVCL) setup with multiple priors, and shows that leveraging multiple experts can alleviate problems in HVCL such as known task boundaries and increased computation cost. To this end, a standard MoE model is built based on  \cite{shazeer2017outrageously} where two different strategies are proposed to encourage expert diversity, which, as the authors believe, may mitigate forgetting since experts specialize in different tasks.

Recently,  pre-trained model based continual learning \cite{zhou2024continual} has attracted much attention by leveraging the powerful pre-trained models to facilitate better continual learning. This further spurs a new category of continual learning approaches, i.e., the prompt-based continual learning approaches \cite{wang2022learning}, which have demonstrated superior performance compared to standard continual learning approaches. Thus motivated, there have emerged several attempts in leveraging MoE to design better prompt-based approaches for continual learning with pretrained models. Specifically,
by shedding lights on the connection between self-attention in Vision Transformer \cite{dosovitskiy2020image} and a mixture of experts, \cite{le2024mixture} interprets applying prefix tuning in pre-trained models as adding new experts, which can be leveraged to facilitate efficient model adaptation for new tasks. Based on this, a novel gating mechanism named Non-linear Residual Gates (NoRGa) is proposed to improve the parameter efficiency in prefix tuning, where non-linear activation functions and residual connections are introduced within the gating function. \cite{li5035279mote} develops a mixture of task-specific experts (MoTE) framework, where a task-specific adapter (i.e., expert) is trained for each new task. To enhance the feature robustness and reduce cross-task feature overlap,
 an expert evaluation strategy and an Expert Peer-Voting Mechanism are further introduced, where the most trustworthy experts will be selected for feature fusion at  inference.

Considering the recent breakthroughs  in large language models, several studies have  explored the usage of MoE in the continual adaptation of language models. \cite{chen2023lifelong} investigates the continual adaptation of pretrained language models, where an MoE-based architecture, i.e., Lifelong-MoE, is proposed to address the forgetting problem. In particular, new experts will be introduced for new data distributions, while regularization-based strategies on both experts and gating functions are used to retain the knowledge for old tasks. A new architecture PMoE is proposed in \cite{jung2024pmoe} to address the forgetting for continual learning in large language models. By using shallow layers for general knowledge and deep layers for new knowledge, new experts are progressively added to deep layers, where a router efficiently allocates new knowledge to appropriate experts based on deep features. 

The application of MoE in continual adaptation based on vision-language models has also been studied very recently. 
\cite{Yu_2024_CVPR} introduces the parameter-efficient MoE-Adapters framework for continual adaptation with the CLIP model \cite{radford2021learning}. This framework leverages Adapters as experts and employs a task-specific router to select the most suitable experts for the current task, thereby accelerating adaptation and fostering expert collaboration. Additionally, the authors propose a Distribution Discriminative Auto-Selector (DDAS) to enhance the model's zero-shot recognition capabilities, which automatically allocates test data to either the MoE-Adapters or the original CLIP model, ensuring accurate predictions for both seen and unseen data. \cite{park2024learning} observes that the usage of a shared layer in MoE-based models can lead to unsatisfying performance for continual learning. To address this, an expert merging strategy is proposed to merge the two most frequently selected experts based on the tracking of expert usage. The merged experts will be used to update the least frequently selected expert. This helps to prevent the same feature being learned as different features by multiple experts.  Additionally, the least important
parts of the model are regularly removed, saving computational
resources.

Moreover, the MoE-based models have  been leveraged in various specific applications of continual learning. For example, 
\cite{zhang2024decomposing} studies the problem of continual test time adaptation and introduces the Mixture-of-Activation-Sparsity-Experts (MoASE) adapter to address the forgetting problem therein.
By decomposing the neuron activation into high and low activation components, MoASE 
develops a Domain-Aware Gate to utilize domain information for expert combination and an Activation Sparsity Gate for more accurate feature decomposition.
\cite{lee2024continual} studies the continual traffic forecasting problem and 
introduces a novel MoE-based model, i.e., TFMoE, to efficiently retain past knowledge. The traffic flow is divided into  homogeneous groups, where each group is handled by an expert model. 
\cite{CL_20} applies MoE to continual medical image segmentation, updating only relevant experts for new data while fixing other network parameters. \cite{li2024theory2} exploits MoE in mobile edge computing (MEC) networks, dynamically selecting experts for task offloading while optimizing gating network parameters to minimize computational delays.

There have also been some studies in the literature that leverage multiple expert models, instead of a single model, for continual learning. For example, 
\cite{aljundi2017expert} proposes a framework where an expert model is trained for each task and a gating mechanism to select the relevant expert at test time. \cite{wang2022coscl} uses a fixed number of small networks to learn all tasks in continual learning in parallel and shows that the performance can be better than a single big model. \cite{doan2023continual} investigates the impact of  different ensemble models on the performance of continual learning. The SEED method proposed in \cite{rypesc2024divide} optimizes the model ensemble training by selecting a single expert for new task learning based on Gaussian distribution similarity, ensuring effective learning of new tasks while preserving prior knowledge.
 EVOLVE \cite{CL_9} leverages multiple pre-trained models as experts to enhance self-supervised continual learning on local clients. In particular, EVOLVE employs an expert aggregation loss to distill guidance information and dynamically adjusts expert weights based on new task data. 
While these studies do not exactly leverage the MoE architecture, they still demonstrate the great potentials of leveraging multiple experts to facilitate better continual learning.

\subsection{Meta-Learning}
The core of meta-learning \cite{sun2019meta,chen2021meta,ren2018meta, santoro2016meta,snell2017prototypical} lies in learning to learn, focusing on enabling models to quickly learn new tasks from a small amount of data, thereby improving the generalization and adaptability of machine learning systems.  MoE can be used to enhance the rapid learning capability of meta-learning methods because it leverages multiple expert models to learn the distinct core features of each domain. By capturing the differences and connections between various tasks in a better way, MoE has great potentials in promoting more efficient learning and adaptation of new tasks by the model. In what follows, we will go over the recent development in the field of meta-learning by leveraging MoE models.

While the most popular meta-learning approach in the past decade is the optimization-based  approach \cite{finn2017model,nichol2018reptile} due to its simplicity and efficiency, neural processes (NPs) based approaches \cite{garnelo2018neural} recently gain increasing attention for meta-learning, which directly learn meta-representations and do not require additional gradient updates during fast adaptation. 
However, the standard NPs consider use a single global Gaussian latentvariable to model tasks, leading to a lack of expressive power when dealing with complex tasks  generated from a mixture of stochastic processes. To alleviate this issue, \cite{wang2022learning} proposes the Mixture of Expert Neural Processes (MoE-NPs), with multiple latent variables and discrete assignment latent variables. The multiple latent variables are used to construct a mixture of NPs experts  to define various functional priors, and the discrete assignment variable controls the expert selection for each data point in prediction. An evidence lower bound is derived to optimize the proposed model, which demonstrates promising performance in both few-shot supervised learning and meta-reinforcement learning.

As meta-learning has demonstrated the superior capability in fast adaptation for novel scenarios, it has been recently applied to generalizable dynamical system reconstruction (DSR) \cite{goring2024out} for learning across varying environments with a small amount of data, which is particular important for scientific discovery using scientific data.
While promising, meta-learning typically relies on the assumption that all tasks (or environments) follow the same task distribution, limiting its applications in scenarios with less similar environments. To address this, \cite{nzoyem2025towards} proposes the Mixture of Experts Reconstructors (MixER) by augmenting existing contextual meta-learning approaches \cite{nzoyem2024extending} with the Top-1 MoE architecture. In particular, MixER consists of multiple experts, each corresponding to a meta-model to capture the shared knowledge, and a linear gating function which takes the context information, instead of state vectors, as input. An unsupervised routing mechanism is further designed to optimize the MoE-based models.

Besides these studies, MoE has also been leveraged together with meta-learning for various applications. Domain adaptation (DA) \cite{farahani2021brief} is a subfield of transfer learning which seeks to adapt the model trained for the source domain to perform well on a related but different target domain. The standard formation of DA considers the transfer from a single domain, whereas multiple sources can be available in practice. 
However, most existing methods use a single model to learn all source tasks. To address this, \cite{guo2018multi} leverages MoE models to explicitly characterize the relationships between different source domains and the target domain, where the prediction for a target example will be based on a weighted combination of all experts' predictions. Here the weights capture the similarity of the target example with each source domain, which are learned through metric-based meta-learning approaches \cite{vinyals2016matching} by constructing a meta-training procedure with multiple multi-source-single-target pairs. \cite{zhong2022meta} investigates a slightly different problem where some unlabeled data from the target domain will be exploited, and propose Meta-Distillation of MoE where a student model is built from each multi-source-single-target pair. Here, a MoE model is used as a teacher student, and each expert is separately trained using supervised learning on each source domain to extract its discriminative features. For domain adaptation, the unlabeled data from the target domain will be passed through all experts and the knowledge from all experts will be aggregated by an aggregator to guide the distillation from the teacher to the student. The student model and the aggregator will be jointly trained using optimization based meta-learning.

Other studies, which also leverage MoE together with meta-learning, investigate some specific applications of domain adaptation. For instance, \cite{MeL_3} studies the domain generalizable person re-identification task (DG ReID),
where each expert in the MoE model learns the knowledge from one source domain and all the features from these experts are integrated based on an adaptive voting process for the unseen target domain. As the specific target domain is generally unknown during training, an optimization-based meta-learning approach is proposed to train the voting network, which is jointly trained with the expert networks given their couplings in the algorithm design. \cite{MeL_6} shares the same design philosophy as \cite{MeL_3} but 
studies the domain generalization of face anti-spoofing, where the domain expert models are meta-trained using optimization-based methods.
 \cite{liu2024meta} investigates the generalization of multi-access control in heterogeneous wireless networks.
Traditional meta-reinforcement learning (meta-RL) methods may not fully capture the subtle differences between different tasks, leading to insufficient adaptability in new environments \cite{rakelly2019efficient}. When designing a General Multiple Access (GMA) protocol, the challenges posed by varying network configurations in different testing environments need to be considered. \cite{liu2024meta} thus introduces the MoE model to enhance the model's representational capacity and reduce overfitting to specific tasks \cite{yuksel2012twenty}. Specifically,  an MoE-enhanced encoder network is introduced in designing the GMA protocol, where each expert model in the linear gating MoE inside the encoder independently encodes each state transition into latent representations. These latent representations are then combined to generate a final comprehensive task representation, which will be leveraged by SAC \cite{haarnoja2018soft} for individual protocol learning within meta-RL of the generalizable GMA protocol learning.

\subsection{Multi-task Learning}

Multi-task learning (MTL) \cite{caruana1997multitask,thung2018brief,zhang2018overview} is a machine learning paradigm whose core idea is to simultaneously learn multiple related tasks to improve the generalization ability and performance of the model. Essentially, it provides the model with rich prior knowledge to enhance its performance. This determines that multi-task learning algorithms need to leverage shared information among tasks, typically achieved by sharing part of the model's parameters or representations. However, this approach has exposed some shortcomings when dealing with complex data \cite{MTL_18}. The MoE model, on the other hand, can intelligently select the most suitable sub-model for processing based on the characteristics of the input data, naturally decoupling different tasks and effectively handling the complex relationships between them. This gives MoE a greater advantage when dealing with complex tasks \cite{gupta2022sparsely}, and as a result, it has gradually gained more attention and been applied to many important sub-tasks in multi-task learning.

To more efficiently handle the relationships among multiple tasks in multi-task learning, \cite{MTL_18} proposes an important variant of MoE called Multi-gate Mixture-of-Experts (MMoE), which sets up a separate gating network for each task. The gating network can further select different expert networks based on the specific characteristics of the current task, decoupling each task into different gate functions. This avoids the mutual interference of tasks with large differences, and alleviates the significant performance degradation of traditional methods when processing multi-task data with large differences due to shared underlying networks. \cite{MTL_21} finds that MMoE still cannot fully resolve the issue of Negative Transfer, and in some tasks, its performance is even worse than that of single-task models \cite{MTL_21}. As a result, they designe several optimization strategies to further mitigate the Negative Transfer problem in MMoE by penalizing overly similar experts, enhancing the fine-grained interaction among experts, and balancing the importance of tasks with varying amounts of data during model training.

Multi-task reinforcement learning is an important branch of multi-task learning \cite{vithayathil2020survey}. It aims to promote the learning of diverse skills by agents to enhance their ability to handle complex tasks. However, existing methods lack guarantees on the diversity of learned representations, which may lead to representation collapse into similar representations, thus limiting their generalization ability \cite{sodhani2021multi,d2024sharing,sun2022paco,devin2017learning,yang2020multi}. To address this issue, \cite{hendawy2023multi} introduces a method called Mixture of Orthogonal Experts (MOOR), which utilizes multiple expert models responsible for generating different representations. In this case, each expert is responsible for different aspects of the task, encouraging the model to learn more diverse skills. To ensure efficient differentiations among experts, MOOR introduces the Gram-Schmidt orthogonalization method into MoE to force each expert to generate mutually orthogonal features, further limiting the possibility of experts generating redundant representations.
\cite{sodhani2021multi} investigates contextual multi-task RL and proposed a novel approach to encode an input observation into multiple representations  using a mixture of encoders. These representations correspond to different skills or objects, and can be selected for any given task based on the context information. This approach
gives the learning agent a more fine-grained control on the information shared across tasks and alleviates negative interference.
Instead of naively sharing parameters across all tasks in traditional approaches,
\cite{cheng2023multi} proposes an attention-based MoE approach for multi-task reinforcement learning, which seeks to learn a compositional policy for each task. In particular, each expert network learns task-specific skills and specializes in different parts of the multi-task representation space.
An attention module was fu introduced to  integrate the  expert outputs, we propose an attention module to generate connections between tasks and experts to achieve the best performance automatically.


Multi-task learning techniques have also garnered significant attention in recommendation systems because they can simultaneously meet the modeling needs of multiple perspectives, improving recommendation performance. However, many existing multi-task recommendation systems struggle to balance parameter sharing and resource utilization. Some MoE-based methods, while more flexible, also face issues such as unstable training or resource waste \cite{MTL_18,jacobs1991adaptive,louizos2017learning}. \cite{MTL_6} improves the standard MoE framework by proposing a new framework called Mixture-of-Masked-Experts (MoME). First, MoME no longer trains independent sub-networks for each expert but extracts expert sub-networks from an over-parameterized base network, generating diverse expert networks by learning different binary masks. Thus, during training, only one base network and a set of binary masks need to be trained, effectively saving resources. Additionally, MoME designs multi-level masks: neuron-level masks are used to filter out unimportant neurons in the base network, while weight-level masks are used to generate diverse expert networks. To further enhance the robustness of the mask learning process, MoME uses a probability formula based on an approximate Bernoulli distribution to determine mask elements and achieves model sparsity through $L_0$ regularization, enabling efficient parameter sharing. \cite{MTL_4} also focuses on recommendation system design, noting that existing multi-task learning models \cite{MTL_18,tang2020progressive} typically use MLP as the expert model. However, using MLP models may not generate distinctive features for different tasks, leading to gradient conflicts and thus affecting model performance. Therefore, this work proposes a new expert network called the Task-Intensive Expert Model, whose parameters are specifically generated by a hyper-network based on different task embeddings, making it more distinctive and alleviating gradient conflict issues. 

\cite{MTL_9} further investigates multi-domain multi-task recommendation tasks, where multi-domain scenarios introduce more complex data dependencies that existing methods cannot handle well \cite{zhang2022leaving}. \cite{MTL_9} decouples the originally complex tasks and introduces shared expert modules, domain expert modules, and task expert modules using MoE. These modules learn cross-domain and cross-task common knowledge, domain-specific user preferences, and task-specific user preferences, respectively , enabling the model to capture user preferences from multiple perspectives and better handle complex dependencies between domains and tasks.

Tasks in machine vision encompass various types, each with its corresponding deep models. However, many of these tasks often follow similar pipeline designs \cite{yang2024multi}. Aggregating the models of these similar tasks into a multi-task model can significantly improve the efficiency of training and inference while maintaining the performance of individual tasks. Considering that MoE not only allows for flexible model expansion but also handles dependencies between different tasks, it is frequently used in the design of visual multi-task models. \cite{MTL_16, fan2022m3vit, MTL_12} primarily apply MoE to improve the encoder. \cite{MTL_16} introduces MoE layers into the attention modules and MLP blocks of ViT \cite{dosovitskiy2020image} and designs a new loss function to encourage strong yet sparse dependencies between tasks and experts, ensuring that each task is only related to a few experts. This can reduce the likelihood of gradient conflicts between tasks. \cite{fan2022m3vit} also incorporates MoE layers into ViT and discusses the advantages and disadvantages of two gating functions in application scenarios: one configures a different router for each task, while the other uses a shared router to achieve dynamic routing by concatenating task embeddings with input tokens. It was found that the first gating method is more effective for handling multiple tasks. \cite{MTL_12} inserts sparsely activated MoE layers into ViT and computes the model's loss on the validation set during each iteration. By using the loss as feedback to dynamically increase or decrease the number of experts,  the cumbersome process of manually adjusting model size can be avoided.

A few works have also attempted to introduce MoE into the decoder. \cite{Ye_2023_ICCV} finds that different task decoders, due to shared decoding parameters, result in a static feature decoding process, leading to insufficient distinction in task-specific features. Therefore, MoE was introduced as a decoder to decompose backbone features into task-agnostic features from different aspects, making the decoding process of task-specific features more fine-grained. This work also utilizes convolutional operations to expand the receptive field of the gating network, which enriches the contextual information in the decision-making process. \cite{yang2024multi} argues that in traditional MoE structures, experts can only interact through routers, thus only establishing connections between some tasks. By adding a 3×3 convolutional layer after the encoder, which is optimized by all tasks, it can learn common features across all tasks, providing the model with more global information to aid decision-making.

The combination of mutli-task learning and MoE has further been explored in other application scenarios.
\cite{MTL_7} replaces the expert networks in MoE with LSTMs \cite{graves2012long}, leveraging the advantages of LSTMs in handling sequential data to improve performance in processing user activity data streams. The work in \cite{MTL_20} applies MMoE to Digital Rock Physics Analysis tasks, addressing the issue of significant computational and memory resource consumption encountered by traditional numerical simulation methods for rock analysis \cite{cui2021vp,karimpouli2018application}. By utilizing MMoE, \cite{MTL_20} transforms serial operations into parallel operations, reducing interference between multiple analysis tasks. \cite{MTL_10} focuses on Heterogeneous Multi-task Learning tasks. In particular, this study only employed the MMoE framework but also introduced some stochastic mechanisms that restrict certain experts to handle specific tasks or disconnect certain experts from the current task, thus enhancing expert exploration.

\subsection{Reinforcement learning}

Reinforcement Learning (RL) plays a pivotal role in decision making in many real-world applications, such as robotics, games, autonomous driving, and healthcare \cite{kober2013reinforcement,szita2012reinforcement,kiran2021deep,yu2021reinforcement}, which seeks to learn a policy by interacting with the environments. However, its practical development suffers from several significant challenges, including computational inefficiency and limited adaptability within high-dimensional state spaces and complex dynamic environments \cite{willi2024mixture}. By dynamically selecting the most appropriate expert based on input data, the MoE mechanism enhances RL agents' flexibility in responding to diverse demand patterns and environmental changes, thus improving overall performance. In what follows, we 
summarize the recent advance in leveraging MoE in RL and 
showcase its potential in various applications. 

The application of MoE in RL has a long history. One notable direction is the so-called modular RL,
which breaks down complex RL problems into smaller modules and builds multiple sub-agents to handle a specific aspect of the task. Notably, \cite{doya2002multiple} considers nonlinear and nonstationary control tasks, and proposes a multiple model-based RL (MMRL) system with multiple modules (i.e., experts). Each module has a dynamics model and a controller, where the state predictions and the action outputs are the weighted combination of all module outputs, with the weights calculated using a gaussian softmax function of the module state prediction error.
Multiple studies have been proposed by following up this work. \cite{samejima2003inter} introduces a concept of `modular reward' which combines both the actual reward and the imaginary reward of appropriate module selections for the task. This will promote independent learning of different modules, which further leads to accurate estimation of global value functions of the composite policy. \cite{van2008switching} constructs multiple experts to specialize in different regions of the overall state space, where standard RL approaches can be used for each expert to capture different information of the underlying state space. This is particularly useful for RL problems with large state spaces.

On the other hand, the idea of modular RL has also been investigated in the design of value-based approaches. Specifically, \cite{peng2016terrain}  introduces a novel mixture of actor-critic experts (MACE) architecture with multiple actor-critic pairs for learning terrain-adaptive dynamic locomotion skills, where each pair will specialize in particular aspects of the motions. MACE is shown to be able to learn more quickly than the standard single actor-critic pair approach. \cite{kumar2017learning} follows this network architecture and learning algorithm to solve MDP problems with a mixture of discrete and continuous actions, in the context of policy learning  for the safe falling problem.
\cite{tommasino2016reinforcement} considers the skill knowledge transfer in multi-task RL and proposes a transfer expert RL architecture (TERL), where both actor and critic consist of a hierarchical structure with the experts and a gating network. The gating network will thus determine the expert selection of different tasks for either policy learning in the actor or value function learning in the critic. Several key features are designed in both gating network and experts to increase the capability of solving new tasks. \cite{peng2019mcp} also considers a multi-task RL framework for transfer learning, which seeks to leverage the pretraining of multiple source tasks to facilitate the learning of new target tasks. Different from the traditional case where only one expert is activated at a particular timestep, this work proposes a model to enable the activation of multiple expert simultaneously. Each expert can specialize in a distribution of actions, whereas the composite policy can be obtained through a composition of these distributions.

More recently, \cite{akrour2021continuous} constructs the policy in policy iteration by leveraging the MoE architecture, where the action of each expert is determined based on the distance to a prototypical state, in order to  ensure the policy is interpretable.
\cite{ren2021probabilistic} introduces Probabilistic Mixture-of-Experts (PMOE), which employs Gaussian mixture models for multi-modal policy representation and frequency approximation gradients to address non-differentiability during optimization. This approach overcomes the limitations of unimodal policies in traditional deep RL, enabling algorithms to learn varied strategies in tasks with multiple optimal solutions. More importantly, PMOE can be potentially applied in generic off-policy and on-policy deep RL algorithms using stochastic policies, e.g., Soft Actor-Critic (SAC) \cite{haarnoja2018soft} and Proximal Policy Optimization (PPO) \cite{schulman2017proximal}.
To improve the sample efficiency of \cite{peng2016terrain} and accelerate model-free RL for terrain-adaptive locomotion skills learning,
\cite{refaat2021accelerated} proposes to generalize  model-based value expansion (MVE) \cite{feinberg2018model}, a technique to obtain better state-action values by using more reliable targets, to a mixture of actor-critic experts. 
\cite{pmlr-v161-gimelfarb21a} proposes an MoE framework incorporating Bayesian functions that autonomously identify and combine promising sub-regions from multiple source tasks across different regions of the state space. By utilizing state-dependent Dirichlet prior distributions to learn similarities between dynamics in source and target tasks, and updating these priors using state transition data from the target environment, this method improves robustness against sparse or delayed rewards, even when there are errors in dynamic estimation.

The attention of leveraging MoE in RL has gained increasing attention last year due to the success of MoE in LLMs.
\cite{obando2024mixtures} investigates of the impact of MoE layers in the DNNs used in value-based deep RL approaches, and shows that the performance of various deep RL algorithms can be significantly improved by incorporating soft MoEs \cite{puigcerver2023sparse} to replay the penultimate layers. Similar phenomenons have been observed in different RL setups, such as online RL, offline RL \cite{levine2020offline}, and RL with a small amount of interactions.
\cite{song2024germ} considers multi-task robot learning and seeks to leverage offline RL to learn from both good demonstrations and sub-optimal data. In particular, the policy network is constructed as an MoE-based Transformer encoder to generate action tokens.
\cite{celik2024acquiring} proposes a Diverse Skill Learning (Di-SkilL) method which aims to enable agents to acquire more skills in RL. In particular, each expert is modeled as a contextual motion primitive, adjusting behavioral strategies according to the current context and optimizing related distributions to focus on optimal performance sub-regions. Di-SkilL effectively addresses hard discontinuities and multimodality in unknown environments by representing each expert's contextual distribution using energy-based models.
\cite{vincze2024smose} proposes a novel method named SMOSE for continuous control tasks. The controller is built upon a Top-1 MoE architecture, where each expert is trained to learn different basic skills and the router is trained to learn to appropriately assign tasks to experts. \cite{willi2024mixture} investigates MoEs' capability to manage non-stationarity — handling and adapting to changes in the environment or data distribution over time — in deep RL settings characterized by ``amplified" non-stationarity via multi-task training. It finds that MoE supports network plasticity and is particularly effective in highly non-stationary conditions, providing fresh insights into addressing non-stationary training environments in RL. 
\cite{huang2024mentor} explores visual deep RL to  enable robots to acquire skills from
visual inputs and proposes a method named MENTOR, which replaces the traditional MLP layers in  visual RL agents by using the MoE architecture for action generations.


The application of MoE in other domains of RL, such as multi-agent RL and imitation learning, has also been explored. For instance, \cite{takahashi2004modular} applies modular RL to soccer robot. \cite{takahashi2005modular} proposes a method to assign multiple modules to different situations, in order to learn purposive behaviors for specified situations related to other agents. Similar to \cite{doya2002multiple}, in \cite{takahashi2005modular} each module consists of a prediction model and a planner, and the module that provides the best estimation of state transitions will be chosen.
\cite{takahashi2005simultaneous} uses modular RL with MoE to learn competitive behaviors in multi-agent systems.
For imitation learning, \cite{mulling2013learning} considers the problem of robot table tennis by allowing robot to learn through interactions with human, where an approach with mixtures of motor experts is proposed to generalize basic movements to various situations. Here each expert corresponds to a motor policy. \cite{ewerton2015learning} constructs a mixture of interaction experts to learn various interaction patterns from  human demonstrations, based on Gaussian mixture models.
\cite{zhou2020movement} leverages Gaussian mixture models to handle the model collapse problem in learning generalized movement representations from human demonstrations. 
\cite{freymuth2022inferring} leverages a mixture of movement primitives to imitate versatile human demonstrations.
\cite{prasad2024moveint} trains a variational autoencoder with a mixture density network to  capture the complexity and variability in Human-Robot Interaction.
Given the diversity of constraints under which demonstration data may be collected, \cite{qiao2023multi} presents Multi-Modal Inverse Constrained Reinforcement Learning (MMICRL), which uses flow-based density estimation for unsupervised expert identification and infers specific agent constraints, thus enhancing the adaptability and reliability of inverse constrained reinforcement learning algorithms.
The strategy of ensemble methods that carries a very similar idea with MoE has also been widely used in RL, in terms of model ensembles \cite{kurutach2018model}, value function ensembles \cite{lee2021sunrise}, and policy ensembles \cite{yang2022towards,lou2023pecan}.



Finally, we summarize several attempts on applying MoE in specific applications of RL. For example, \cite{gupta2023offline} confronts the challenge of driving chatbots with RL by developing several RL algorithms tailored for Dialogue Management. Leveraging the hierarchical structure of MoE-based  language models enables offline RL methods to operate within a significantly reduced action space, making RL problems more tractable while generating discourse that reflects diverse intentions. Similarly, in video processing \cite{mohammadi2023reinforcement} and investment portfolio management \cite{RL_21}, MoE-based approaches dynamically activate specific expert models based on current data, achieving higher accuracy, lower computational costs, and enhanced adaptability to changing environments.
\cite{li2024mixtures} leverage MoE models to scale up the parameters of RL models in optimal execution tasks.






\subsection{Federated learning}

Federated Learning \cite{mcmahan2016federated} is proposed to address the challenge of how different data owners can effectively exchange data without compromising user privacy and ensuring data security. By first training multiple local models on different servers and then aggregating these local model parameters using various rules to update the global model, the federated learning paradigm not only fully leverages distributed yet information-rich data but also efficiently utilizes the computational resources of multiple clients, inspiring a series of subsequent work, particularly in the field of recommendation systems \cite{liu2022federated, wu2021fedgnn,flanagan2020federated}.

When updating the global model using multiple local model parameters, one of the most common rules is Federated Averaging (FedAVG) \cite{mcmahan2017communication}. This rule assigns weights to the corresponding local  models based on the size of each client's local dataset during parameter updates, and then updates the global model using the weighted combinations of local parameters. However, this rule relies on a manually defined simple heuristic, which not only requires each model to be initialized with the same random seed for convergence but is also unsuitable for scenarios where the data distribution among clients is highly heterogeneous \cite{mcmahan2017communication}. Considering the hybrid nature of MoE, \cite{peterson2019private} is the first to introduce the MoE architecture into the federated learning setting. Specifically, this work proposes to train a general model using aggregated data and domain-specific models for each client using private data, and then use a gating mechanism to determine the contribution of each model's output to the learning task. Compared to FedAVG, this approach better generates outputs based on task characteristics. \cite{zec2020specialized} adopts a similar approach but focuses more on scenarios with stronger data heterogeneity across clients. Additionally,  an opt-out mechanism is designed to ensure that clients with sensitive data could choose not to participate in the data integration process, thereby safeguarding data security. Some other work has also followed this combined approach \cite{pye2021personalised}.

An important goal of federated learning is to effectively integrate data from multiple clients. However, it is evident that real-world data distributions are highly complex and variable. Therefore, a major challenge in federated learning is data heterogeneity, which can manifest in two ways: one where clients share the same conditional distribution but have different marginal distributions, and the other where the opposite is true. Both scenarios can harm the model's learning performance. Recently some studies have made substantial efforts to address this challenge. For example, \cite{reisser2021federated} introduces two improvements: First, this work incorporates client identifier information into the gating decision; Second, to reduce the difficulty of training MoE expert models on non-IID data, which could lead to model degradation, a variational lower bound is leveraged to derive a unified global variational approximation, which replaces the true posterior distribution of all clients during training. \cite{guo2021pfl} proposes an improved model fine-tuning method, where a general model is first trained and each client only fine-tunes the shallow parameters in the general model. Additionally, for complex high-dimensional data, this work uses partial model parameters to extract features from the raw data before feeding them into the gating function to make inputs distinguishable. \cite{isaksson2022adaptive} first trains local models on each client and then uses the clustering method IFCA \cite{ghosh2020efficient} to iteratively cluster all clients, generating a global cluster model for each cluster. Finally,  a gating mechanism is used to combine the outputs of the global cluster models and the local models to produce the final result. \cite{tran2025revisiting} also introduces a MoE layer on the server side but dynamically determines the number of top-K experts to retain based on the client's resource conditions during routing, fully utilizing limited computational resources.

In addition to the discussion on heterogeneity, another significant challenge in federated learning is communication overhead. Each round of communication between decentralized servers incurs substantial costs. Generally, there are two approaches to address this issue. The first approach is to reduce the amount of information that needs to be communicated in each round by transmitting only critical model parameters. For example, the number of experts in the locally trained MoE layer can be reduced, discarding some expert parameters that contribute less to the output in exchange for improved communication efficiency \cite{dun2023fedjets,tran2025revisiting}. Alternatively, only the most important MoE models can be selected for transmission. The second approach is to reduce the number of communication rounds, such as in One-shot Federated Learning (OFL), where typically only a single round of communication is performed during training to aggregate client models on the central server. After this, clients no longer participate in subsequent optimization updates. However, many popular OFL methods \cite{heinbaugh2023data,su2023one} employ lossy aggregation, where aggregating all local models into a single global model results in knowledge loss, especially when model heterogeneity is strong \cite{guha2019one}. In contrast, \cite{zeng2024one} proposes to treat all local models as expert models and train a specialized lightweight gating function using additional data to aggregate these experts. This approach effectively preserves the knowledge of local models while reducing communication overhead.

\section{Theory}\label{sec:sec_4}

Although the MoE model has a long history and achieved great empirical success in practice during the past few years \cite{eigen2013learning,shazeer2017outrageously}, the theory behind it is largely under-explored, especially in deep neural networks. Recently, some studies have initiated the attempt to build theoretical understandings of MoE. In this section,  we will delve into these studies, aiming to provide the readers a basic idea of the theory development in this area.

Most of the theory development for MoE so far mainly focus on simple models. Early studies in this area seek to understand the approximation capacity of MoE models. For instance, 
\cite{jiang1999hierarchical} investigates the convergence rate for hierarchical MoE models where the experts are exponential family regression models. \cite{jiang1999approximation} characterizes the convergence rate for hierarchical MoE models with generalized linear models as the experts. \cite{zeevi1998error} provides the error bounds for functional approximation using MoE models and shows the convergence to any sufficiently differentiable target function in the Sobolev space.
\cite{nguyen2016universal} provides an alternative result of this through a universal approximation theorem for MoE models with softmax gating functions and linear experts, without imposing assumptions on the differentiability of the target function. 
\cite{mendes2012convergence} considers a standard MoE model with softmax gating functions where the experts are polynomial regression models, and characterizes the convergence rate of the maximum likelihood estimator (MLE) depending on the number of experts and the order of the polynomial model.

Recently, some studies have emerged and aimed to characterize the convergence of MLE for MoE models with different gating functions for regression  problems.  For instance, \cite{ho2022convergence} characterizes the convergence rate of the MLE for Gaussian MoE models with covariate-free gating functions, by leveraging techniques from optimal transport. \cite{nguyen2023demystifying} attempts to understand the parameter estimation of the MLE for  Gaussian MoE models with softmax gating functions,  By leveraging  novel Voronoi loss functions to capture the interactions between the gating and experts,  the convergence rate of MLE is established. The convergence of MLE for Gaussian MoE models with other types of gating functions, such as Gaussian density gating \cite{nguyen2023statistical} and top-K sparse softmax gating \cite{nguyen2024towards}, has also been investigated. \cite{nguyen2024least} studies a more general setup of MoE models with softmax gating for least square estimation, and characterizes the convergence rates for different types of expert model.  
The theoretical understanding behind the recently emerged cosine gating function is developed in \cite{nguyen2024statistical} for lease square estimation.

In contrast, the theoretical understanding of MoE models for classification problems is less explored. \cite{nguyen2023general} investigates the classification problem with the MoE models including softmax gating and multinominal logistic expert models. In particular, the convergence rates for density estimation and parameter estimation are established. A novel class of modified softmax gating functions is further proposed to improve the convergence rate for parameter estimation when part of expert parameters vanish. \cite{nguyen2023general} also studies the inefficiency issues encountered by the gating function in MoE during training. Specifically, when some expert parameters vanish (i.e., the parameters of certain expert models approach zero or become insignificant), the interaction between the standard softmax gating function and the expert functions through partial differential equations (PDEs) significantly slows down the parameter estimation rate \cite{ho2022convergence,nguyen2024towards,nguyen2023demystifying,nguyen2023statistical}. This paper thoroughly investigated the problem under an improved softmax gating function setting. Specifically, they used a bounded function $M(X)$ to transform the input $X$ before passing it to the softmax gating function. Although the expert parameters still depend on the input $X$ , the transformed $X $ no longer has a linear relationship with the original $M(X)$, thereby eliminating the interaction between gating parameters and expert parameters and significantly improving the model's convergence speed and stability.

Most of the successes of MoE models nowadays happen with more complex expert models, such as deep neural networks. However, the theoretical understanding of MoEs in deep learning remains elusive compared to the theoretical studies of MoE in simple models, largely due to the limited understanding of deep neural networks. \cite{chen2022towards} seeks to fill this gap and makes the first attempt to understand the behaviors of MoEs in deep learning. More specifically, they consider a standard MoE architecture with softmax gating, where each expert model is a two-layer CNN, and investigate a binary classification problem where the dataset contains multiple clusters. The authors show that any single-expert model with a two-layer CNN cannot achieve more than 87.5\% accuracy on this specific dataset, while a linear MoE model performs slightly better than a single-expert model but still falls short of a nonlinear MoE model. With sufficient expert exploration during the training, it can be shown that 
the router can automatically learn the cluster structure of the data and dynamically route the data to the most suitable expert. Following this work, 
\cite{chowdhury2023patch} investigates the patch-level routing for MoE models (pMoE) with linear gating in solving a supervised binary classification task, where each expert is a two-layer CNN. They show that the pMoE model can achieve similar generalization performance with standard CNN but with a reduced sample complexity. One of the key reasons behind this, also theoretically and empirically justified, is that an appropriately trained patch-level router can route the class-discriminative patches of one class to the same expert and drop some class-irrelevant patches.

With the increasingly broad applications of MoE in many problems, the theoretical studies of MoE start to emerge beyond the general machine learning setup. \cite{fung2025mixture} explores the usage of MoE models in handling multilevel data while building the theoretical understanding therein. Notably, multilevel data  is very common in practical applications \cite{aitkin1986statistical,goldstein1986multilevel}, such as  hierarchical data with a nested structure where different levels are correlated. Ignoring the dependencies within the data can lead to spurious, misleading, or biased clustering and prediction outcomes \cite{goldstein2011multilevel}. To better capture the  dependencies, \cite{fung2025mixture} proposes a Mixed MoE to handle multi-level data, which we define as MeMoE in this work. This approach not only introduces random effects into the gating function, making the computed gating values depend not only on input variables but also on random effects. This helps the model better capture random effects in multi-level data when dynamically adjusting expert weights. At the same time, to simplify the model, the expert functions are separated from the random effects, and the probability distribution of the expert model no longer depends on input variables but only on output variables. Under certain regularity conditions, this paper proves that the MeMoE is dense in the sense of weak convergence for any continuous mixed-effects model. In the special case of hierarchical multi-level data, this paper further proves that MeMoE can approximate the complex dependency structures of random effects between different factor levels in multi-level data.

\cite{li2024theory1} investigates how integrating the MoE framework into models can effectively mitigate generalization and forgetting issues in continual learning \cite{mccloskey1989catastrophic,kirkpatrick2017overcoming}, under the setup of over-parameterized linear regression tasks and a linear gating MoE model. The authors prove that a sufficiently trained MoE model can diversify experts to specialize in different tasks, which minimize the interference across tasks and reduce the forgetting in continual learning. To ensure the stability of the gating function, the update of the gating network must be terminated in a timely manner for new task learning. Explicit expressions for expected forgetting and overall generalization error are also derived, quantitatively evaluating the impact of MoE in continual learning. \cite{li2024theory2} theoretically investigates the usage of the MoE framework in mobile edge computing (MEC)  on handling continuous task streams therein. Specifically, each MEC service station is treated as an expert model, updating its local model based on the data distribution of tasks. An adaptive gating network was proposed to dynamically allocate tasks to different experts, ensuring that each expert converges and specializes in specific task types. Additionally, the paper derived the minimum number of experts required to ensure system convergence and proved that the MoE model can control the overall generalization error within a small constant range, significantly outperforming traditional MEC offloading strategies.

\section{Applications} \label{sec:sec_5}

In this section, we investigate the applications of MoE in two broad application domains, computer vision (CV) and natural language processing (NLP). By examining the influence of MoE in tasks ranging from image classification to machine translation, we highlight how this architectures enhance specialization, scalability, and adaptability, offering insights into their practical impact and future potential across diverse real-world scenarios.

\subsection{Computer vision}

With the rapid development and maturation of the CV field, the focus of both academia and industry has gradually shifted towards how CV technologies can be effectively implemented in specific application scenarios. However, this process has encountered several challenges such as resource constraints in edge deployment, difficulties in model scalability, and computational efficiency. These issues have pointed to the need of the redesign of existing models, especially when dealing with large volumes of data and complex tasks, where traditional model architectures often struggle to find an ideal balance between performance, flexibility, and stability. Since MoE has attracted renewed attention due to its performance advantages demonstrated in Mixtral 8x7B~\cite{jiang2024mixtral}, an increasing number of studies have attempted to apply it in the CV field, to boost the state-of-the-art performance. Since the MoE architecture introduces multiple specialized experts to handle specific types of inputs and dynamically adjusts the activation levels of different experts based on the input data, it not only helps alleviate hardware resource bottlenecks but also enhances the model's ability to handle diverse visual tasks, thereby optimizing both performance and stability. Therefore, in this subsection we  will focus on the application of MoE in the field of CV by reviewing recent advancements from four bedrock aspects, i.e., classification, detection, segmentation, and generation. The aim is to provide valuable references for future explorations to address the challenges in the CV field.

\textbf{1) Image Classification.}
Image classification, as one of the most fundamental machine learning tasks, is a key component of many CV tasks. Therefore, understanding the performance and limitations of MoE in image classification can, to some extent, reflect its potentials for application in other CV tasks.

Notably, \cite{riquelme2021scaling} proposes and validates a new approach, namely V-MoE, to cope with visual tasks, by replacing the MLP layers in Vision-Transformer \cite{dosovitskiy2020image} with sparse MoE layers. V-MoE can  scale the model to 15 billion parameters  and demonstrates the efficiency gains from this simple combination in image classification tasks.
Further explorations within the traditional MoE framework have been conducted. \cite{videau2024mixture} dissects the MoE architecture, providing a detailed exploration of its training and optimization. The paper discusses how the number of expert networks, the number of MOE layers, and their placement affect the model's performance, efficiency, and stability. 

As expert networks are a key component of MoE, many efforts have been made to optimize them. ViMOE~\cite{han2024vimoe} introduces the concept of shared experts, where a shared expert handles common knowledge required for classification, while other specialized experts focus on specific knowledge. This approach mitigates the difficulty in exploring optimal configurations due to the sensitivity of MoE layers to expert setups and enhances training stability. The paper also systematically analyzes routing behaviors, examining routing strategies, the number of experts, and the placement of MoE layers. It visualizes routing behaviors with heatmaps, finding that MoE layers closer to the output handle more semantically rich feature maps, leading to more specialized tasks for experts. Consequently, the last L layers of the model should be replaced with sparse MoE layers. The analysis suggests that both the number of experts and the size of L are related to the degree of routing, and more does not necessarily mean better performance. To some degree, the base model introduced in \cite{royer2023revisiting}  shares similarities with the shared expert concept proposed in ViMOE~\cite{han2024vimoe}. In CLIP-MoE \cite{zhang2024clip}, the MoE architecture is applied to enhance the capabilities of the CLIP model through a method called Diversified Multiplet Upcycling. The authors fine-tune a series of pre-trained CLIP models using multi-stage contrastive learning, extracting expert networks that focus on different aspects of the input data, and integrating these experts into the MoE architecture.

Other papers delve deeper into the design of gating functions. \cite{puigcerver2023sparse} introduces a novel technique called Soft MoE, which differs from traditional hard assignment of input tokens to experts by using a weighted average of all tokens for soft assignment. Therefore, each expert processes different parts of these weighted combinations. This method retains the advantages of the MoE architecture in scaling model capacity. Besides, it also improves training stability, reduces inference time, and effectively addresses the token dropping issue common in MoE training, creating a new paradigm for gating functions. ~\cite{royer2023revisiting} addresses the overfitting problem encountered when using MOE on small datasets by designing an architecture that incorporates early exit mechanisms and pre-training. Specifically, the model pre-trains a base model during the training phase and pre-determines the number of experts using k-means clustering. During inference, if the base model has high confidence in a sample's prediction, it can opt for an early exit, bypassing the subsequent MoE layers for specialized learning. Otherwise, the sample is passed to the MoE layers for specialized learning, and the outputs from both the MoE layers and the base model are combined by Ensemblers to produce the final output.

~\cite{nguyen2024expert} focuses on improving the design of gating mechanisms in Hierarchical Mixture of Experts (HMoE) models to address performance bottlenecks in handling complex inputs and executing specific tasks. Specifically, this work proposes a method that goes beyond the traditional Softmax gating function by using Laplace gating strategies for both hierarchical and task allocation. By customizing gating functions for each expert group, the HMoE model can allocate resources more effectively without increasing computational burden and improve performance on complex datasets. 
Additionally, it explores the intrinsic interactions between first-level and second-level gating parameters, which significantly affect the convergence speed of model parameter estimation, providing a thoughtful direction for allocating complex or hierarchical datasets. \cite{he_deepme_2021} designs a deep mixture algorithm called DeepME (Deep Mixture Experts) for efficiently handling large-scale image processing tasks. It groups similar image categories based on semantic relevance, allowing for some overlap between categories. This ensures that each task group contains image categories with similar learning complexities, and specific base deep networks are trained for these groups. Additionally, a gating network is trained to combine all base deep networks, generating a mixed network with larger outputs to effectively handle large-scale datasets containing thousands of categories.

\textbf{2) Object Detection.}
Object detection is also one of the important tasks in the field of CV. Building on image classification, it  requires the model to  not only detect objects but also  output the location of each classified object, where ensuring  the computational efficiency of the model is more challenging compared to  that in image classification.

Similarly, many studies have explored the impact of simply integrating MoE with base models for object detection. MoCaE~\cite{oksuz2023mocae} has shown that directly using simple methods from deep ensemble (DEs) \cite{lakshminarayanan2017simple} to integrate different object detectors with the MoE architecture does not improve model performance and may even have adverse effects. 
The underlying reason is that 
simply adding different detectors leads to unfair competition, thereby affecting the final detection results. To address this, this work introduces Early Calibration and Late Calibration to adjust the confidence levels of different detectors, better reflecting their true detection performance. Experimental results have shown that MoCaE yields significant gains over single models and DEs on several real-world challenging detection tasks.

~\cite{wang2024object} replaces the FFN layer in the traditional transformer with the MoE-HCO block, proposing an event-stream-based object detection framework called MvHeat-DET. The MoE-HCO block selects the most suitable transformation branch for the current features through a policy network, providing various signal transformation expert networks. Therefore,  the input events can have more diverse and suitable processing modules. It also uses frequency embedding to predict thermal diffusion coefficients to simulate the heat conduction process. This will enable the model to better capture spatiotemporal dynamic features in event streams while maintaining efficient computation, demonstrating high practical values. To help the model better understand and process multi-source datasets, DAMEX~\cite{jain2024damex} proposes a data-aware MoE architecture, by replacing the FFN layer in the transformer with an MoE layer. Different experts are trained to learn data from different sources, enhancing the model's generalization capability. Traditional two-stage methods \cite{zhou2022simple,wang2019towards}  require multiple classification heads to detect data from different datasets, leading to an increase in parameters with mixed datasets. In contrast, the data-aware MoE layer can achieve excellent performance without significantly increasing the number of parameters. \cite{feng2024pluralistic} addresses the diversity and ambiguity of the task itself, by transforming the original single-mask prediction task into a multi-mask prediction task and predicting human preference scores for each possible salient object mask. To enable the model to handle multiple tasks more efficiently, this paper leverages the MoE architecture to solve the integrated training problem of the two tasks, and overcomes the inconsistency in input and output formats between the two subtasks. In particular, this work  replaces the FFN layer in DaViT \cite{ding2022davit} with an MoE layer, which contains two experts: the P-FFN layer for handling the PSOD (Pluralistic Salient Object Detection) task and the Q-FFN for handling the MQP (Mask Quality Predictor) task. This allows the model to dynamically select the most suitable expert network for the current task, improving model performance without significantly increasing the number of parameters.

\textbf{3) Semantic Segmentation.}
Semantic segmentation requires classifying each pixel in the input image. As a result, processing high-resolution images may significantly increase the demand on the hardware, hindering the model's practical applications. 
Compared to traditional segmentation models, the MoE-based model can provide the following benefits: 1)
The MoE architecture alleviates hardware pressure through its sparse activation and divide-and-conquer strategy, enhancing the model's practicality. 2) Some studies \cite{pavlitska2024towards,zhu2024customize} have also found that using MoE can achieve excellent performance in semantic segmentation tasks that require greater model stability and generalization. 3) MoE also enjoys a better interpretability in terms of knowledge specialization.

~\cite{wang2020deep} seeks to leverage MoE to achieve a good balance between computational complexity and representational capacity in deep neural networks. In particular,  traditional convolutional networks are combined with shallow embedding networks and multi-head sparse gating networks, dynamically selecting and executing partial networks at each convolutional layer. This can improve the accuracy while reducing prediction costs without significantly increasing the network width. Furthermore, this work proposes two DeepMoE variants including wide-DeepMoE and narrow-DeepMoE. The first one is suitable for applications requiring high accuracy, whereas the second one is designed for computational resource-limited scenarios. Swin2-MoSE~\cite{rossi2025swin2} demonstrates that MoE can also be applied to improve semantic segmentation models, especially on remote sensing images. The authors design an MoE layer called MoE-SM, which includes an SM (Smart Merger) module to merge the outputs of various experts and adopts a new per-example strategy instead of the commonly used per-token one. This can ensure that all tokens of each example are processed by the same expert, enhancing the model's competitiveness in semantic segmentation tasks.

Some studies have also investigated the usage of MoE in the segmentation for autonomous driving related scenarios. For example, \cite{pavlitskaya_using_2020} aims to leverage MoE to better understand autonomous driving scenarios which can further lead to improved algorithm design. Due to its architecture design, MoE inherently enjoys better transparency. Compared to other general models \cite{lakshminarayanan2017simple,kendall2017uncertainties}, the MoE models can not only provide the final overall model output, but also enable the analysis of the outputs of individual expert models and their consistency or divergence with the overall output during the decision-making process. This mechanism provides more detailed information about how the models work, offering better interpretability while maintaining performance close to that of a single baseline model.
A common issue in segmenting urban and highway traffic scenes is vulnerability to adversarial attacks. 
To address this, 
\cite{pavlitska2024towards} attempts to use MoE given its properties of more dynamic model selections compared to traditional methods \cite{abbasi2017robustness,kariyappa1901improving,pang2019improving}. This work shows that the MoE models exhibits higher robustness against instance-specific attacks, universal white-box adversarial attacks, and cross-model transfer attacks, maintaining relatively high accuracy under these attacks. Additionally, the MoE models with additional convolutional layers show even stronger resistance to attacks. These results indicate that MoE not only improves model performance but also enhances model robustness and stability.

\textbf{4) Image Generation.}
Image generation can produce realistic and diverse images, which is very useful in many applications \cite{rombach2022high,ledig2017photo,wei2018person,radford2015unsupervised}. However, current image generation technologies still face various challenges such as low generation quality, limited diversity, and poor adaptability to complex tasks. By decomposing the complex problem of image generation into multiple simpler and core tasks, MoE leverages the collaboration among the experts to enhance the quality of generated images, improve the diversity of outputs, and further enrich detail representations.

RAPHAEL~\cite{xue2023raphael} introduces two types of MoE models  for temporal control and spatial settings during generation, respectively. On one hand, a spatial MoE layer is responsible for depicting different text concepts in specific image regions, allowing each text token to learn specific visual features through specialized experts. This can enhance the representation of different concepts. On the other hand, a temporal MoE layer focuses on processing these concepts at different time steps of the diffusion process to handle varying degrees of noise impact. This configuration results in billions of diffusion paths from the network input to the output, each path acting as a ``painter" for specific concepts and image regions, achieving finer text-to-image alignment and improving the quality and aesthetic appeal of generated images. The application of MoE  enables RAPHAEL to flexibly switch between multiple styles and closely adhere to text prompts.

GANs \cite{goodfellow2020generative} are known to struggle to learn multimodal data distributions when processing complex datasets, leading to  generated images with poor-quality. To address this, \cite{park2018megan}  proposes MEGAN with multiple generator networks, where each network focuses on learning specific modal distributions in the dataset. This design allows each generator to focus on different data subsets, generating more diverse and higher-quality images.

A novel design of MoA is introduced in \cite{wang2024moa} by creating two attention paths, i.e., a personalized branch and a non-personalized prior branch, allowing the model to retain its powerful generation capabilities while minimizing interference with the personalized part. This specially designed MoA can intelligently adjust the generation process based on the input text and subject image. By optimizing the fusion of personalized and generic content,   high-fidelity personalized image generation can be achieved without sacrificing image diversity or quality. Additionally, the learning routing mechanism in \cite{wang2024moa} dynamically manages pixel distribution at each level, further improving the performance for image generation tasks that involve multiple subjects and complex interactions.

Text2Human~\cite{jiang2022text2human} applies MoE to the transformer encoder based on the diffusion model \cite{gu2022vector,esser2021imagebart,bond2022unleashing}, enabling conditional generation of clothing with different texture types. Specifically, the MoE model routes input features to different expert heads based on the texture attributes mentioned in the text prompt, with each expert head responsible for predicting tokens of specific textures. Based on the synthesis and control of more complex textures, this method increases the model's ability to handle details while avoiding the computational burden of training separate samplers for each texture. 

\subsection{Natural Language Processing}

As a core research area of artificial intelligence, NLP  focuses on enabling computers to understand, generate, and process human language \cite{manning1999foundations,jurafsky2000speech}. It plays a significant role in promoting human-computer interaction, information extraction, knowledge mining, cross-lingual communication, and automation. However, the field of NLP still faces challenges such as the trade-off between model capacity and computational cost, poor adaptability to multi-task and multi-domain scenarios, sparse data and long-tail problems, insufficient reasoning and logical capabilities, and the need for personalization and dynamic adaptability. 

The MoE architecture effectively alleviates these challenges through mechanisms such as sparse activation and the allocation of different expert models for different tasks or domains. This not only reduces computational costs but also enhances model performance in multi-task, low-resource scenarios, and complex reasoning tasks, while supporting personalized services. 
In light of the application of MoE in the NLP field, this subsection will summarize and review relevant work from the perspectives of Natural Language Understanding (NLU) and Natural Language Generation (NLG).

\textbf{1) Natural Language Understanding.} 
The advancement of NLU technology enables machines to understand and interpret human language, 
by bridging the gap between human communication and machine processing \cite{hirschberg2015advances,young2018recent}. This allows systems to
perform tasks such as intent recognition, entity extraction, and semantic analysis \cite{devlin2019bert,liu2019roberta}, 
which are crucial for applications such as virtual assistants, chatbots \cite{adiwardana2020towards}, sentiment analysis \cite{zhang2018deep}, and information extraction \cite{lample2016neural}.  However, the further development of NLU is hindered by the complexity, ambiguity, and diversity of human language. MoE allows systems to dynamically leverage expert networks for different linguistic tasks or domains. For example, one expert may specialize in syntactic parsing, while another focuses on sentiment analysis, thereby enhancing the model's ability to effectively handle various complex linguistic patterns. This specialization and adaptability make MoE a powerful method for extending and refining NLU systems, enabling more accurate and context-aware language understanding. Since the use of specially designed MoE architectures to enhance model performance in NLG tasks has not yet reached a scale sufficient for sub-domain classification, this section will only summarize some representative work in recent years.

GLaM \cite{du2022glam} introduces an MoE architecture to make the training and inference processes more efficient. It significantly reduces the required computational resources and energy consumption while maintaining or even improving model performance. This encouraging result highlights the great potentials in leveraging MoE to address the issue of high computational resource consumption in large-scale, intensive language models for NLU tasks.

MoE-LPR \cite{zhou2024moe} designs a two-phase training strategy for the MoE-based model. In the first phase, the original model is converted into an MoE architecture, and new expert modules are added, which improves the model's capability for new languages without using original language data. In the second phase, a small amount of original language data is used for review, and a language prior routing mechanism is employed to restore and maintain the performance of the original language. Experimental results demonstrate the excellent scalability and stability of MoE-LPR, providing a new perspective for multilingual natural language understanding tasks.

When speech is transcribed into text, errors from Automatic Speech Recognition (ASR) systems \cite{yu2016automatic} often affect the accuracy of subsequent NLU components. To address this, \cite{cheng_moe-slu_2024} proposes an MoE-based framework, namely MoE-SLU, aiming to mitigate the impact of ASR errors on spoken language understanding performance. MoE-SLU employs three strategies to generate additional transcripts and uses MoE  to weight and average these transcripts. By enhancing the model's ability to capture keywords, MoE-SLU is more robust to ASR errors,  achieving state-of-the-art performance on three benchmark SLU datasets. The performance of MoE-SLU can be further improved through regularized predictions.

To address the issue of interference in multi-task learning, particularly among NLU tasks, MT-TaG \cite{gupta2022sparsely} introduces a sparsely activated MoE architecture and designs a task-aware gating mechanism to route inputs to specific expert networks. Compared to traditional dense models, MT-TaG demonstrates superior performance in multi-task learning, especially in low-resource task transfer, efficient generalization, and handling unrelated tasks. MoPE-BAF \cite{wu2024mixture} designs various soft prompt experts, including text prompts, image prompts, and unified prompts, based on a unified vision-language model to enrich unimodal representations and promote multimodal interaction. By introducing a block-aware prompt fusion mechanism, the model achieves cross-modal prompt attention across Transformer layers, smoothly transitioning from unimodal representations to multimodal fusion. Experimental results show that MoPE-BAF significantly improves performance in few-shot settings for multimodal sarcasm detection and sentiment analysis tasks, demonstrating the effectiveness of MoE in deep multimodal semantic understanding tasks.

\textbf{2) Natural Language Generation.}
NLG can transform structured data or concepts into natural and fluent text, widely applied in machine translation \cite{bahdanau2014neural}, dialogue systems \cite{wen2015semantically}, content creation \cite{reiter1997building}, and information summarization \cite{rush2015neural}, among other fields. To achieve diverse and precise language expression, NLG systems require strong model capacity to capture complex linguistic structures and contextual dependencies, while effectively handling interference in multi-task learning to maintain high performance across different scenarios. MoE allows models to dynamically select the most suitable expert networks and flexibly adjust model capacity, reducing task interference and enhancing model specialization, thereby contributing to the construction of more powerful and flexible NLG systems.

\emph{(a) Text Generation.} Text generation has numerous applications in NLP. Through effective text processing, NLP systems can accurately understand, generate, and transform human language, which is crucial for achieving more advanced NLP applications. However, text processing still faces challenges such as insufficient generation diversity and uncontrollable generated content. MoE dynamically selects the most suitable sub-network for specific tasks, making models more efficient and precise when processing complex or diverse text data. 

In \cite{10095401}, MoE is introduced into the generator of a language GAN, leveraging multiple experts to collaboratively generate high-quality sentences, with each expert acting as a recurrent neural network that autoregressively produces the current token representation. Additionally, the Feature Statistics Alignment (FSA) paradigm is incorporated to further optimize the learning signals during generator training, making the generated text more closely aligned with the real data distribution. In RetGen \cite{zhang2022retgen}, during inference, a retriever first obtains the top K most relevant documents and their corresponding probabilities. Then, K independently trained transformer-based generation models process each document along with the same context, combining the partially generated results with the current consensus to produce their respective output distributions. The final output distribution is a weighted combination of these independent generation results, with weights determined by the document relevance probabilities. This approach overcomes issues such as ignoring document relevance information when simply concatenating multiple documents as input, allowing the model to more effectively utilize document relevance information to guide the text generation process. LogicMoE \cite{wu_enhancing_2024} focuses on solving the problem of table-to-text generation. In the designed LogicMoE architecture, each expert acts as a specialized generator for a specific logical type, responsible for generating sentences that meet the requirements of that logical type. This design not only improves the quality of generated sentences but also enriches the diversity of generated content at the semantic and logical levels. Experimental results show that LogicMoE achieves absolute improvements of 0.8 and 2.2 points in BLEU-3 scores \cite{papineni2002bleu}, demonstrating the inherent advantages of LogicMoE in generation diversity and controllability.
The optimization of MoE system design has also been explored
\cite{frantar2023qmoe}  to enhance model efficiency in text processing tasks.

\emph{(b) Machine Translation.} Machine translation is the application of automatically translating text from one language to another, greatly facilitating cross-lingual communication, information access, and global collaboration \cite{brown1990statistical,wu2016google}. However, machine translation faces challenges such as linguistic diversity, grammatical complexity, and cultural differences. MoE uses conditional computation to dynamically allocate tasks, enabling the system to select the most suitable expert based on the input content. This mechanism not only improves translation accuracy and fluency but also better handles complex scenarios involving multiple languages and domains, thereby enhancing the performance and adaptability of machine translation.

\cite{shazeer2017outrageously} innovatively introduces Sparsely-Gated MoE technology for conditional computation. Applying MoE to language modeling and machine translation tasks not only significantly improves model performance on large datasets but also achieves efficient utilization of computational resources. Gshard \cite{lepikhin2020gshard} replaces traditional feedforward network layers with MoE layers and adopts conditional computation to achieve efficient utilization of computational resources and flexible expansion of model capacity. MoE-based  Transformer models at different scales are designed and trained to meet translation needs from high-resource to low-resource languages. Experimental results show that the MoE  models trained using GShard   significantly improve translation quality, while achieving higher training efficiency and lower costs compared to dense models under the same hardware conditions. \cite{costa2022no} also introduces a selective activation mechanism by replacing some dense model layers with MoE layers, significantly enhancing the model's representational capacity with similar inference and training efficiency. This approach is particularly beneficial for high-resource languages due to the increased model capacity, while also helping low-resource languages reduce interference from unrelated languages. Additionally, to address the issues in large-scale MoE models, such as quick overfitting in low-resource directions, regularization and curriculum learning strategies are applied to optimize complex training dynamics. A good balance is achieved between cross-lingual transfer and interference in multilingual machine translation. \cite{nllb_team_scaling_2024} also leverages sparse activation mechanisms to significantly enhance the model's representational capacity without increasing computational costs.

In order to address the inefficiency issues encountered during the deployment and inference stages of MoE models,
\cite{huang2023towards} proposes  three optimization techniques, including dynamic gating, expert caching, and load balancing methods, to improve the inference efficiency of MoE models in language modeling and machine translation tasks.

\textbf{3) Multimodal Fusion.} Multimodal fusion integrates information from text and other modalities, providing richer and more comprehensive data representations. In particular, multimodal fusion can capture contextual clues and semantic details that single modalities cannot provide, thereby enhancing the model's understanding and predictive capabilities. In multimodal fusion scenarios, using MoE allows for dynamic adjustment of the importance weights of different modalities based on their characteristics, enabling effective integration and utilization of information from different sources while avoiding the high complexity and optimization difficulties faced by traditional single models. This not only improves the overall performance of the model but also enhances its flexibility and scalability, allowing for better performance in complex multimodal tasks.

LIMoE \cite{mustafa2022multimodal}, as the first large-scale multimodal mixture of experts model, can simultaneously process image and text data and align their representations through contrastive learning. Additionally, this study demonstrates the organic emergence of modality-specific experts in LIMoE, where experts spontaneously specialize in handling specific types of data or tasks,. This indicates that the model not only effectively allocates resources for different tasks but also promotes cross-modal and cross-task knowledge transfer.
LLaVA-MoLE \cite{chen2024llava} addresses the issue of data conflicts in multimodal large language models when fine-tuning with mixed-domain instruction data, by proposing the use of sparse LoRA Mixture of Experts (MoLE). The proposed method selects the most suitable LoRA expert for each input token based on its embedding features in the attention layers, allowing the model to adapt to inputs from different domains. This  effectively mitigates performance degradation caused by mixing different types of instruction data.  The Hunyuan model \cite{sun2024hunyuan} is the first Chinese MoE multimodal large model, achieving a score of 71.95 on SuperClue-V, surpassing some international top models and demonstrating its unique advantages in the context of Chinese culture.

\section{Future Directions} \label{sec:sec_6}

MoE models present significant opportunities for advancing machine learning in real-world applications, but they also face several challenges that must be addressed to unlock their full potentials. In this section, we discuss some of the key future research directions for MoE models.

\subsection{Training stability and load balancing}

One of the most critical challenges in MoE models is ensuring training stability and load balancing among experts. Due to the dynamic nature of expert selection, some experts may receive significantly more data than others, leading to imbalanced training and potential model collapse. Future work should focus on developing more robust training strategies that ensure balanced utilization of experts. Techniques such as adaptive load balancing, dynamic expert capacity adjustment, and regularization methods that penalize over-reliance on specific experts could be explored. In some cases, theoretical studies on the convergence properties of MoE models under different load balancing strategies would provide valuable insights into designing more stable training algorithms. Additionally, the sparsity of MoE during training also poses challenges for stable model training, requiring more work to explore and optimize existing training strategies.

\subsection{Training and system efficiency}

While MoE models offer the advantage of conditional computation, their training and inference efficiency remain a concern, especially in large-scale applications. Hardware advancements have reduced computational costs, but high latency remains a persistent issue. Future research should focus on optimizing hardware-software co-design, particularly for conditional computation. Techniques such as efficient memory management, reduced communication overhead, and parallel processing strategies could significantly improve the scalability of MoE models. Combining MoE with conditional computation techniques may also open new avenues for dynamic resource allocation and task-specific adaptations, further enhancing efficiency. 


\subsection{Architecture design}

The design of MoE architectures, particularly the determination of the number of experts and their specialization, is another area that requires further exploration. Current approaches often rely on heuristic methods or static setting to decide the number of experts, which may not be optimal for all tasks. Future work should investigate more principled approaches to determining the number of experts, such as using meta-learning or reinforcement learning to dynamically adjust the architecture based on task complexity and data distribution. Additionally, novel architecture designs that integrate MoE with other neural network components, such as attention mechanisms or graph neural networks, could lead to more powerful and flexible models.

\subsection{Theory development}

Improving the interpretability of MoE models remains a critical area for future work. While MoE architectures have demonstrated remarkable scalability and performance, the theoretical underpinnings of their behavior—such as expert routing decisions and clustering mechanisms—are not yet fully understood, especially in modern deep neural networks. More rigorous theoretical studies are needed to explain these characteristics, which could lead to more robust and reliable models. This would also shed light on the design of better gating function and expert network as well. 

\subsection{Tailored algorithm design}

MoE models have shown great promise in various machine learning paradigms, but their potential in combination with other learning paradigms remains underexplored. Future work should investigate the integration of MoE with other learning frameworks, such as contrastive learning, transfer learning, and self-supervised learning. For example, in federated learning, MoE models could be used to handle heterogeneous data distributions across clients by assigning different experts to different clients. Exploring these combinations could lead to more versatile and powerful learning systems.

\subsection{New applications}

Although MoE has been extensively explored in NLP, its potential in other domains remains underexplored. For instance, in computer vision, recent work has shown that MoE models can achieve performance that continues to improve with training, suggesting untapped potential in areas like image segmentation, object detection, and multimodal learning. Further exploration in these domains could yield significant breakthroughs. Additionally, MoE models could be applied to emerging fields such as healthcare, robotics, autonomous systems, education, finance, as well as recommendation systems, where the ability to handle diverse and complex tasks is crucial. Developing MoE models tailored to these specific applications could lead to more effective and efficient solutions.

\section{Conclusion} \label{sec:sec_7}

This comprehensive survey delves into the integration of the MoE architecture with various domains.
Starting from various basic designs of the MoE architecture and training strategies, we 
highlight its synergy with important machine learning algorithms, alongside the recent theoretical advancements for understanding MoE. Furthermore, we provide a systematic summary of MoE in two critical application domains, computer vision and natural language processing, and shed light on the important future designs for improving the design and impact of MoE.
We hope that this work will serve as a valuable reference for 
researchers in the field and beyond, 
elicit escalating attentions, and inspire further research ideas in MoE.

\bibliographystyle{ieeetr}
\bibliography{main}

\end{document}